\documentclass{article}

\usepackage{arxiv}

\usepackage[utf8]{inputenc} % allow utf-8 input
\usepackage[T1]{fontenc}    % use 8-bit T1 fonts
\usepackage{hyperref}       % hyperlinks
\usepackage{url}            % simple URL typesetting
\usepackage{booktabs}       % professional-quality tables
\usepackage{amsfonts}       % blackboard math symbols
\usepackage{nicefrac}       % compact symbols for 1/2, etc.
\usepackage{microtype}      % microtypography
\usepackage{lipsum}
\usepackage{rotating}
\usepackage{subcaption}
\usepackage{lscape}
\usepackage{graphicx}
\usepackage{xcolor}
\graphicspath{ {./images/} }

\title{A combined full-reference image quality assessment approach based on convolutional activation maps}

\author{Domonkos Varga}

\begin{document}
\maketitle
\begin{abstract}
The goal of full-reference image quality assessment (FR-IQA) is to predict
the perceptual quality of an image as perceived by human observers using its
pristine (distortion free) reference~counterpart.
In this study, we explore a novel, combined approach which predicts the
perceptual quality of a distorted image by compiling a feature vector from
convolutional activation maps. More specifically, a reference-distorted image
pair is run through a pretrained convolutional neural network and the
activation maps are compared with a traditional image similarity metric.
Subsequently, the resulting feature vector is mapped onto perceptual
quality scores with the help of a trained support vector regressor.
A detailed parameter study
is also presented in which the design choices of the proposed method is explained.
Furthermore, we study the relationship between the amount of
training images and the prediction performance. Specifically,~it~is demonstrated
that the proposed method can be trained with a small amount of data
to reach high prediction performance.
Our~best proposal---called ActMapFeat---is compared to the state-of-the-art on
six publicly available benchmark IQA databases,
such as KADID-10k, TID2013, TID2008, MDID, CSIQ, and VCL-FER. 
Specifically, our method is able to
significantly outperform the state-of-the-art on these benchmark~databases.
\footnote{This manuscript is a preprint version of "Varga, D. A Combined Full-Reference Image
Quality Assessment Method Based on Convolutional
Activation Maps. \textit{Algorithms} \textbf{2020}, 13, 313." }
\end{abstract}

% Keywords

\section{Introduction}
\label{sec:intro}
In recent decades, a continuous growth in
the number of digital images
has been observed, due to the spread of smart phones and various social media.
As a result of the huge number of imaging sensors,
there is a massive amount of visual data being produced each day.
However, digital images may suffer different distortions during
the procedure of acquisition, transmission, or compression. As a result,
unsatisfactory perceived visual quality or a certain level of
annoyance may occur. Consequently, it is essential to predict the perceptual
quality of images in many applications, such as display technology, communication,
image compression, image restoration, image retrieval,
object detection, or image registration.
Broadly speaking,
image quality assessment (IQA) algorithms can be classified into three different
classes based on the availability of the reference, undistorted image. 
Full-reference (FR)
and reduced-reference (RR) IQA algorithms have full and partial
information about the reference image, respectively. In contrast, no-reference
(NR) IQA methods do not posses any information about the reference image.

Convolutional neural networks (CNN), introduced
by LeCun {et al.} \cite{lecun1998gradient} in 1988,
are used in many applications, from 
image classification \cite{szegedy2015going} to audio 
synthesis \cite{oord2016wavenet}. In 2012, 
Krizhevsky {et al.} \cite{krizhevsky2012imagenet} won the
ImageNet \cite{russakovsky2015imagenet}
challenge by training a deep CNN relying on
graphical processing units. Due to the huge number of
parameters in a CNN, the training set has to
contain sufficient data to
avoid over-fitting. However, the number of human annotated images in many
databases is rather limited to training a CNN from scratch.
On the other hand, a CNN trained on
ImageNet database \cite{russakovsky2015imagenet} is able to provide
powerful features for a wide range of image processing 
tasks \cite{sharif2014cnn,penatti2015deep,bousetouane2015off},
due to the learned comprehensive set of features.
In this paper, we propose
a combined FR-IQA metric based on the comparison
of feature maps extracted
from pretrained CNNs. The rest of this section is
organized as follows.
In Section~\ref{sec:related}, previous and related
work are summarized and
reviewed. Next, Section \ref{sec:contributions} outlines the
main contributions of this study.

\subsection{Related Work}
\label{sec:related}
Over the past few decades, many FR-IQA algorithms have been proposed
in the literature.
The~earliest algorithms, such as mean squared error (MSE) and
peak signal-to-noise ratio (PSNR), are~based on the energy of image
distortions to measure perceptual image quality.
Later, methods~have appeared that
utilized certain characteristics of the human visual system (HVS).
This kind of FR-IQA algorithms can be classified into two groups:
{bottom-up} and {top-down} ones.
Bottom-up approaches directly build on
the properties of HVS, such as
luminance adaptation \cite{chou1995perceptually},
contrast sensitivity \cite{daly1992visible},
or~contrast masking \cite{watson1997image}, to create a model
that enables the prediction of perceptual quality.
In~contrast, top-down methods try to incorporate the general characteristics
of HVS into a metric to devise effective algorithms. Probably, the most famous
top-down approach is the structural similarity index (SSIM) proposed by
Wang {et al.} \cite{wang2004image}.
The main idea behind SSIM \cite{wang2004image}
is to make a distinction between structural and
non-structural image distortions, because the HVS is mainly sensitive to the latter
ones. Specifically,~SSIM is determined at each coordinate within local
windows of the distorted and the reference images. The distorted image's overall
quality is the arithmetic mean of the local windows' values.
Later, advanced forms of SSIM have been
proposed. 
For example, edge-based structural similarity \cite{chen2006edge} (ESSIM) compares the edge
information between the reference image block and the distorted one, claiming
that edge information is the most important image structure information for the HVS.
MS-SSIM \cite{wang2003multiscale} built multi-scale
information to SSIM, while 3-SSIM \cite{li2009three} is
a weighted average of different SSIMs for edges, textures, and smooth regions.
Furthermore, saliency weighted \cite{liu2011visual} and information content
weighted \cite{wang2010information} SSIMs were also introduced in the
literature.
Feature similarity index (FSIM) \cite{zhang2011fsim} relies on
the fact that the HVS utilizes
low-level features, such as edges and zero crossings,
in the early stage of visual information
processing to interpret images. This is why FSIM utilizes
two features: (1) phase
congruency, which is a contrast-invariant dimensionless
measure of the local structure and
(2) an image gradient magnitude feature.
Gradient~magnitude similarity deviation (GMSD) \cite{xue2013gradient} 
method utilizes the sensitivity of
image gradients to image distortions and pixel-wise gradient similarity combined with
a pooling strategy, applied for the prediction of the perceptual image quality.
In contrast, Haar wavelet-based perceptual similarity 
index (HaarPSI)~\cite{reisenhofer2018haar} applies coefficients obtained
from Haar wavelet decomposition to compile an IQA metric.
Specifically, the magnitudes of high-frequency coefficients were used to
define local similarities, while the low-frequency ones were applied to weight
the importance of image regions.
Quaternion image processing provides a true vectorial
approach to image quality assessment.
Wang {et al.} \cite{wang2008color} gave a quaternion 
description for the structural information
of color images. Namely, the local variance of the luminance was taken as the real
part of a quaternion, while the three RGB channels were taken as the
imaginary parts of a quaternion. Moreover, the perceptual quality was
characterized by the angle computed between the singular value feature vectors of
the quaternion matrices derived from the distorted and the reference image.
In contrast, Kolaman and Pecht \cite{kolaman2011quaternion} created a
quaternion-based structural similarity index (QSSIM) to assess the quality of RGB
images. A study on the effect of image features, such as contrast, blur, granularity,
geometry distortion, noise, and color, on the perceived image quality can be found
in \cite{glowacz2010automated}.

Following the success in image classification \cite{krizhevsky2012imagenet},
deep learning has become extremely
popular in the field of image processing.
Liang {et al.} \cite{liang2016image} first introduced
a dual-path convolutional neural network (CNN)
containing two channels of inputs. Specifically, one input channel was dedicated
to the reference image and another for the distorted image. Moreover, the presented
network had one output that predicted the
image quality score. First, the input distorted
and reference images were decomposed into $224\times224$-sized image patches and the
quality of each image pair was predicted independently of each other. Finally, the
overall image quality was determined by averaging the scores of the image pairs.
Kim and Lee \cite{kim2017deep} introduced a similar dual-path CNN but their model
accepts a distorted image and an error map calculated from the reference and the
distorted image as inputs. Furthermore, it generates a visual sensitivity map which is
multiplied by an error map to predict perceptual image quality. Similarly to the
previous algorithm, the inputs are also decomposed into
smaller image patches and the overall
image quality is determined by the averaging of the scores of
distorted patch-error map pairs. 

Recently, generic features extracted from different pretrained CNNs, such as
AlexNet \cite{krizhevsky2012imagenet} or GoogLeNet \cite{szegedy2015going}, 
have been proven
very powerful for a wide range of
image processing tasks.
\mbox{Razavian {et al.} \cite{sharif2014cnn}}
applied feature vectors
extracted from the OverFeat \cite{sermanet2013overfeat} network, which was
trained for object classification
on ImageNet ILSVRC 2013 \cite{russakovsky2015imagenet}, 
to carry out image classification, 
scene recognition, fine-grained
recognition, attribute detection, and content-based image retrieval. The~authors
reported on superior results compared to those of traditional algorithms.
Later, Zhang~{et~al.}~\cite{zhang2018unreasonable} pointed out that
feature vectors extracted from pretrained CNNs outperform traditional image
quality metrics. Motivated by the above-mentioned results, a number of
FR-IQA algorithms have been proposed relying on different deep features and
pretrained CNNs.
\mbox{Amirshahi {et al.} \cite{ali2017image}} compared different
activation maps
of the reference and the distorted image extracted from
AlexNet \cite{krizhevsky2012imagenet}
CNN. Specifically,~the~similarity of the activation maps was measured to
produce quality sub-scores. Finally,~these sub-scores were aggregated
to produce an overall quality value of the distorted image.
In~contrast, Bosse {et al.} \cite{bosse2017deep}
extracted deep features with the help of
a VGG16 \cite{simonyan2014very} network from reference and
distorted image patches. Subsequently, the distorted and the reference deep
feature vectors were fused together and mapped onto patch-wise quality scores.
Finally, the patch-wise scores were pooled, supplementing with
a patch weight estimation procedure to obtain the overall perceptual
quality. In our previous work \cite{varga2020composition}, we introduced a composition
preserving deep architecture for FR-IQA relying on a Siamese layout of pretrained CNNs,
feature pooling, and a feedforward neural~network.

Another line of works focuses on creating combined metrics where
existing FR-IQA algorithms are combined to achieve strong correlation with
the subjective ground-truth scores.~In \cite{okarma2010combined}, Okarma~examined the properties of three FR-IQA metrics
(MS-SSIM \cite{wang2003multiscale}, VIF \cite{sheikh2006image}, and
R-SVD \cite{mansouri2009image}), and
proposed a combined quality metric based on the
arithmetical product and power of these metrics.
Later, this approach was further developed using optimization
techniques \cite{okarma2012combined,okarma2013extended}.
Similarly,~Oszust~\cite{oszust2017image} selected
16 FR-IQA metrics and applied their scores as
predictor variables in a lasso regression model to obtain a combined metric.
Yuan {et al.} \cite{yuan2015image} took
a similar approach, but kernel ridge
regression was utilized to fuse the scores of the IQA metrics.
In contrast, Lukin {et al.} \cite{lukin2015combining} fused the results
of six metrics with the help of a neural network. Oszust \cite{oszust2016full}
carried out a decision fusion based on 16 FR-IQA measures by minimizing
the root mean square error of prediction performance with a genetic~algorithm.

\subsection{Contributions}
\label{sec:contributions}
Motivated by recent convolutional activation map
based metrics \cite{ali2017image,amirshahi2018reviving}, we make the following contributions in our study.
Previous activation map-based approaches compared directly the similarity between reference and distorted activation maps by
histogram-based similarity metrics. Subsequently,~the~resulted sub-scores were pooled together using different ad-hoc solutions, such
as geometric mean.
In contrast, we take a machine learning approach. Specifically, we compile a feature vector for each distorted-reference image
pair by comparing distorted and reference activation maps with the help of traditional image similarity metrics. Subsequently, these feature
vectors are mapped to perceptual quality scores using machine learning techniques.
Unlike previous combined \mbox{methods \cite{okarma2012combined,okarma2013extended,lukin2015combining,oszust2016full},} we do not apply directly different
optimization or machine learning techniques using the results of traditional
metrics; instead, traditional metrics are used to compare convolutional activation
map and to compile a feature vector.

We demonstrate that our approach has several advantages.
First, the proposed FR-IQA algorithm can be easily generalized to any
input image resolution or base CNN architecture, since image patches are
not required to crop from the
input images like several previous
CNN-based \mbox{approaches \cite{bosse2017deep,liang2016image,kim2017deep}.} In this regard, it is similar to recently
published NR-IQA algorithms, such as DeepFL-IQA \cite{lin2020deepfl}
and BLINDER \cite{gao2018blind}.
Second, the proposed feature extraction method is highly
effective, since the proposed method is able to reach state-of-the-art results even if only 5\%
of the \mbox{KADID-10k \cite{lin2019kadid}} database is used for training.
In contrast, state-of-the-art deep learning based approaches' performances are strongly
dependent on the training database size \cite{lin2018koniq}. Another advantage of the proposed approach is that it is able to achieve the performance of traditional FR-IQA metrics, even in cross-database tests. 
Our method is compared against the state-of-the-art on six publicly available IQA benchmark databases,
such as KADID-10k \cite{lin2019kadid}, TID2013 \cite{ponomarenko2009tid2008}, VCL-FER \cite{zaric2012vcl}, MDID \cite{sun2017mdid}, 
CSIQ \cite{larson2010most},
and TID2008 \cite{ponomarenko2009tid2008}.
Specifically,~our~method is able to
significantly outperform the state-of-the-art on the benchmark~databases.
%Different from previous
%works \cite{bosse2017deep,liang2016image,kim2017deep}
%where the image features were extracted from one
%or two layers of a pretrained CNN, we exploit deep
%features of multiple stages
%and combine the deep features of the
%reference and distorted images to
%effectively predict the distorted
%images' perceptual quality.
%Namely, a CNN constructs a
%hierarchical representation of input images.
%Deeper layers contain higher-level features,
%while earlier layers encode low-level features.
%We show that incorporating both low-level and high-level deep representation
%into a regression model results in improved performance.
\subsection{Structure}
The remainder of this paper is organized as follows.
After this introduction, Section \ref{sec:proposed} presents
our proposed approach.
Section \ref{sec:experimental} shows experimental results and analysis 
with a parameter study, a~comparison to other
state-of-the-art methods, and a cross-database test.
Finally, a conclusion is drawn in Section \ref{sec:conc}.

\section{Proposed Method}
\label{sec:proposed}
The proposed approach is based on constructing feature vectors from each
convolutional layer of a pretrained CNN for a reference-distorted
image pair. Subsequently, the convolutional layer-wise feature
vectors are concatenated and mapped onto perceptual quality scores
with the help of a regression algorithm.
In our experiments, we used the AlexNet \cite{krizhevsky2012imagenet}
pretrained CNN
which won the 2012 ILSVRC by reducing the error rate
from 26.2\% to 15.2\%. This was the first time that a CNN performed so well
on ImageNet database. The techniques, which were introduced
in this model, are~widely used also today, such as data augmentation
and drop-out. In total, it contains five convolutional and three
fully connected layers. Furthermore, rectified linear unit (ReLU)
was applied after each convolutional and fully connected layer
as activation function.

\subsection{Architecture}
In this subsection, the proposed deep FR-IQA framework, which aims to capture
image features in variuous levels from a pretrained CNN,
is introduced in details. 
Existing works extract features of one or two layers from
a pretrained CNN in FR-IQA \cite{bosse2017deep,varga2020composition}.
However, many papers pointed out the advantages of
considering the features of multiple layers in NR-IQA \cite{gao2018blind} and
aesthetics quality assessment \cite{hii2017multigap}.

\begin{figure}
\centering
\includegraphics[width=10cm]{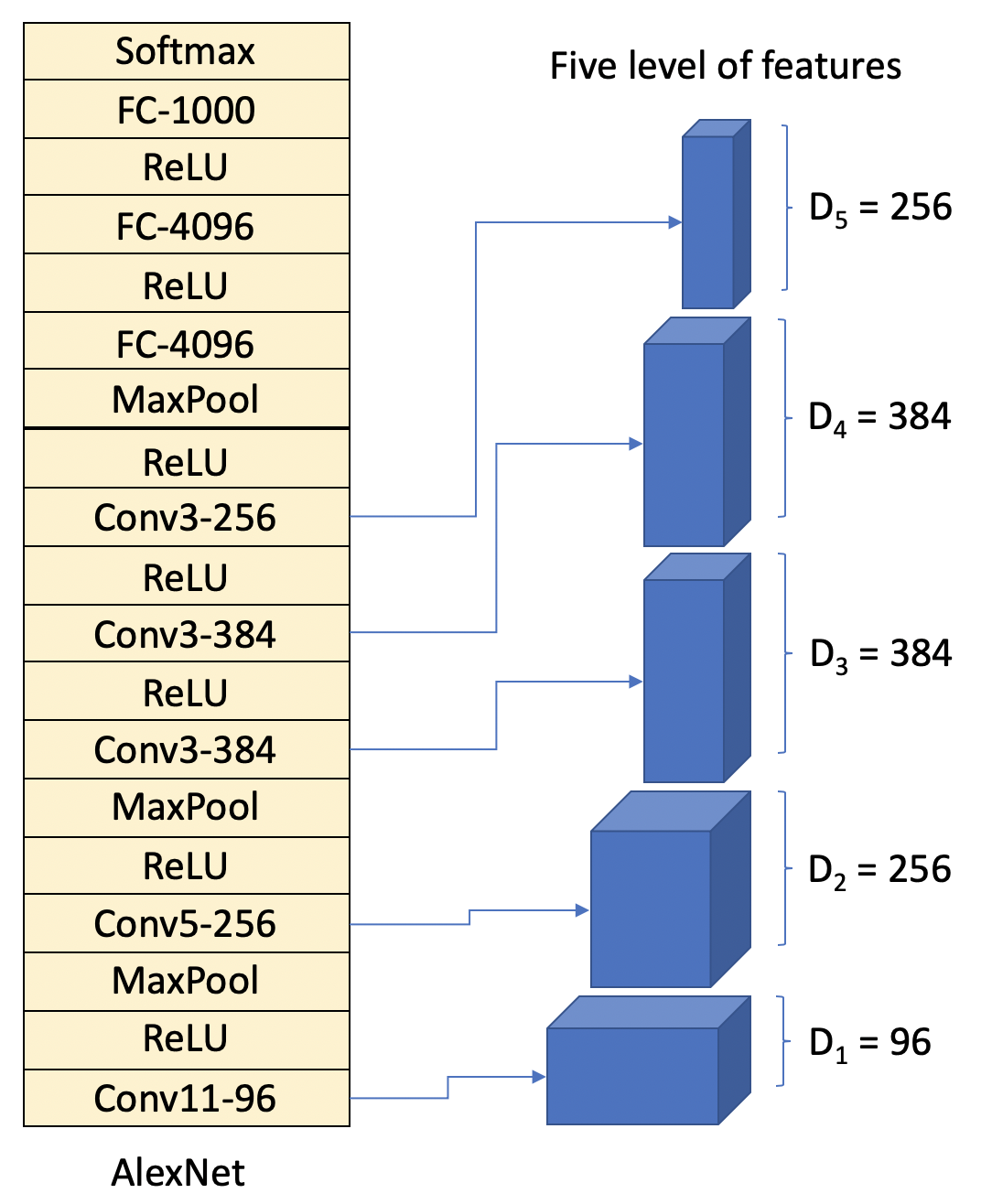}
\caption{Feature extraction with the help of AlexNet \cite{krizhevsky2012imagenet}.}
\label{fig:5LevelOfFeatures}
\end{figure}

\begin{figure}
\centering
\includegraphics[width=13.5cm]{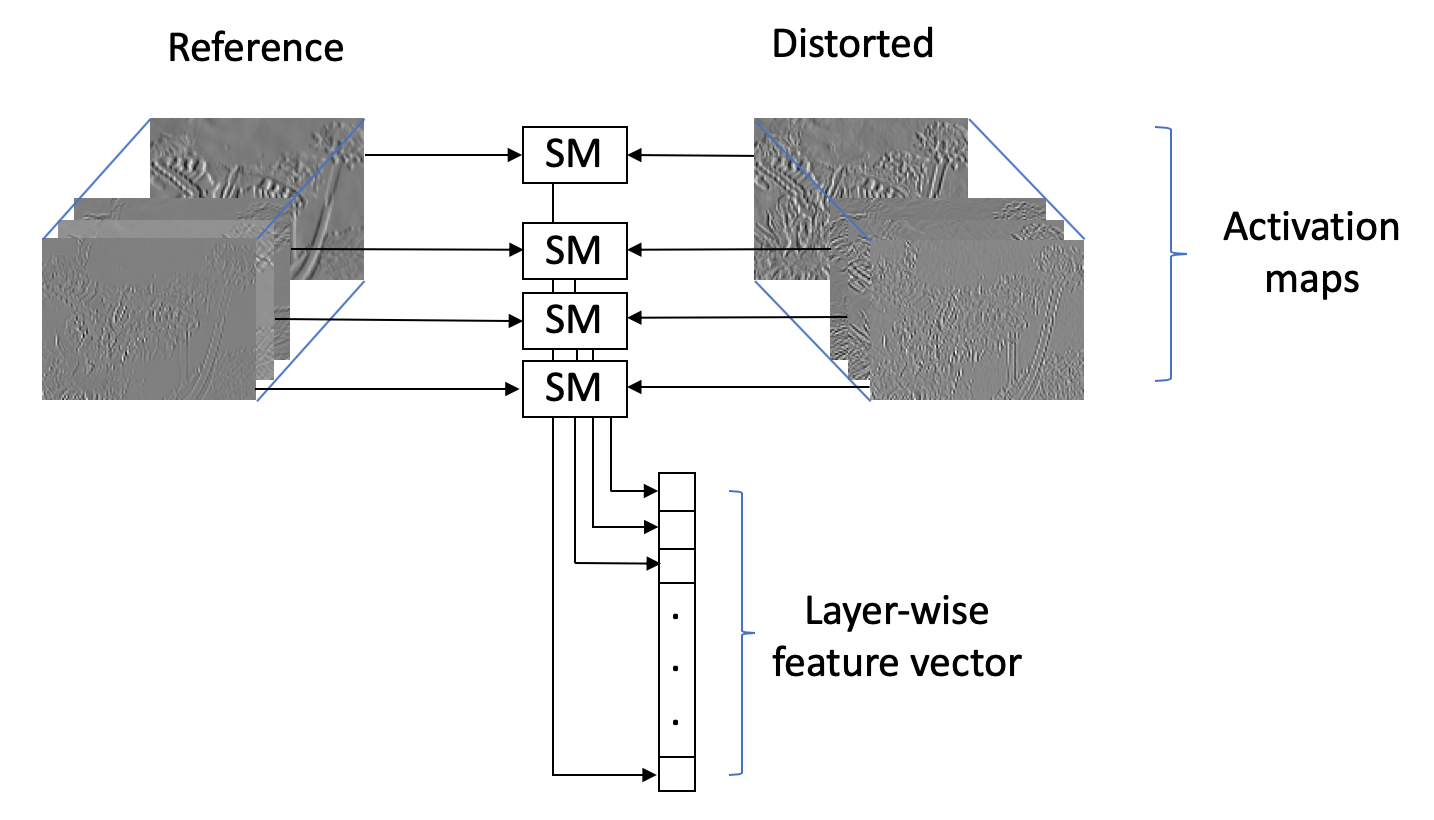}
\caption{Compilation of layer-wise feature vectors. The activation maps of a distorted-reference image pair are compared to each other in a given
convolutional layer with the help of traditional image similarity metrics (SM).}
\label{fig:layerwise}
\end{figure}

We put the applied
base CNN architectures into a unified framework by slicing a CNN into
$L$ parts by the convolutional layers, independent from the network
architecture, {e.g.}, AlexNet or VGG16.
Without the loss of generality, the slicing of AlexNet \cite{krizhevsky2012imagenet} is 
shown in Figure \ref{fig:5LevelOfFeatures}.
As one can see from Figure \ref{fig:5LevelOfFeatures},
at this point, the features are in the form of $W\times H\times D$
tensors, where the depth $(D)$ is dependent on the applied base CNN
architecture and the tensors's width $(W)$ and height $(H)$
depend on the input image size.
In order to make the feature vectors' dimension independent
from the input image pairs' resolution, we do the followings.
First, convolutional feature tensors are extracted
with the help of a pretrained CNN (Figure \ref{fig:5LevelOfFeatures})
from the reference image and from the corresponding distorted
image.
Second, reference and distorted 
activation maps at a given convolutional layer
are compared using traditional image similarity metrics.
More specifically, the $i$th element of a layer-wise
feature vector corresponds to the similarity between the
$i$th activation map of the reference feature tensor and
$i$th activation map of the distorted feature tensor.
Formally, we can write
\begin{equation}
\textbf{f}^l_i = ISM(\mathcal{F}^{ref,l}_i, \mathcal{F}^{dist,l}_i) 
\end{equation}
where $ISM(\cdot)$ denotes a traditional image similarity
metric (PSNR, SSIM \cite{wang2004image}, and
HaarPSI \cite{reisenhofer2018haar}
are considered in this study), 
$\mathcal{F}^{ref,l}_i$ and $\mathcal{F}^{dist,l}_i$ are the
$i$th activation map from the $l$th reference and distorted
feature tensors, $\textbf{f}^l$ is the feature vector extracted
from the $l$th convolutional layer, and $\textbf{f}^l_i$ stands for
its $i$th element. Figure \ref{fig:layerwise} illustrates the compilation of layer-wise
feature vectors.

\begin{figure}
    \centering
    \begin{subfigure}{0.475\textwidth}
        \centering
        \includegraphics[height=1.5in]{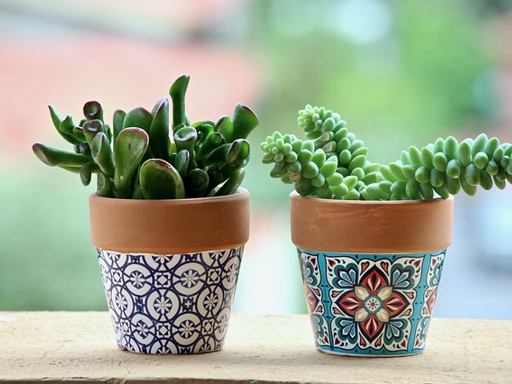}
        \caption{}
        \label{fig:imgsRef}
    \end{subfigure}%
    ~ 
    \begin{subfigure}{0.475\textwidth}
        \centering
        \includegraphics[height=1.5in]{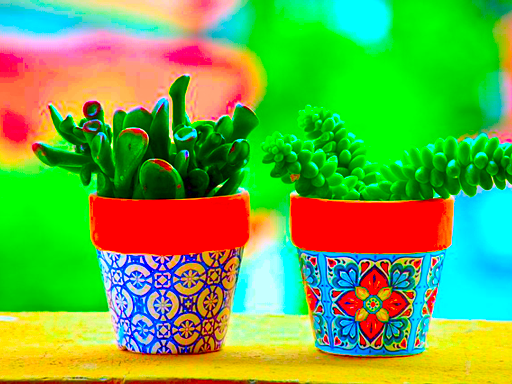}
        \caption{}
        \label{fig:imgsDist}
    \end{subfigure}
    \caption{A reference-distorted image pair from KADID-10k \cite{lin2019kadid}. (\textbf{a}) Reference image. (\textbf{b})~Distorted~image.}
\label{fig:imgs}%MDPI: Explanations moved to caption, please confirm. 
                %AUTHOR: Confirmed.  
\end{figure}

\begin{figure}   
    \centering
    \begin{subfigure}{0.475\textwidth}
        \centering
        \includegraphics[height=1.5in]{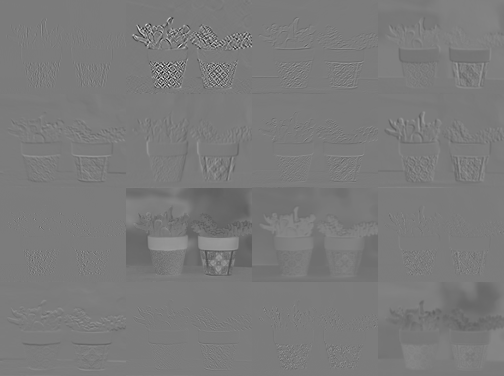}
        \caption{}
    \end{subfigure}
    ~
    \centering
    \begin{subfigure}{0.475\textwidth}
        \centering
        \includegraphics[height=1.5in]{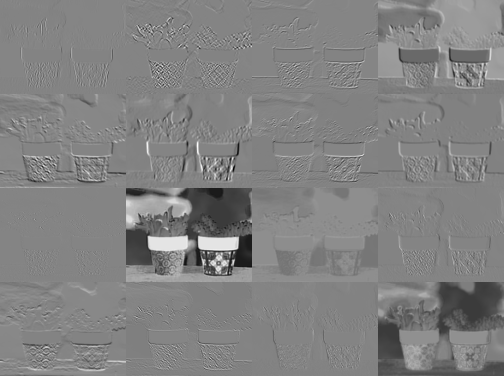}
        \caption{}
    \end{subfigure}%
    \caption{Activation map visualization of a reference-distorted image pair. (\textbf{a}) Visualization of the first 16 activation maps of AlexNet's \cite{krizhevsky2012imagenet} \textit{conv1} layer using the reference image in Figure \ref{fig:imgs}a. (\textbf{b}) Visualization of the first 16 activation maps of AlexNet's \cite{krizhevsky2012imagenet} \textit{conv1} layer using the distorted image in Figure \ref{fig:imgs}b.}
    \label{fig:maps}
\end{figure}

In contrast to other machine learning techniques, CNNs are often
called black-box techniques due to millions of parameters and
highly nonlinear internal representations of the data. Activation~maps of an input image
help us to understand which features that a CNN has learned. If we feed AlexNet reference and distorted image pairs
and we visualize the activations of the \textit{conv1} layer, it can be seen that activations of the reference image
and those of the distorted image differs significantly from each other mainly in quality aware details, such as edges
and textures (see Figures \ref{fig:imgs} and \ref{fig:maps} for illustration). This observation revealed to us that an effective feature vector can be compiled by comparing
the activation maps.

The whole feature vector, that characterizes a 
reference-distorted image pair, can be obtained by concatenating
the layer-wise feature vectors. Formally, we can write
\begin{equation}
\textbf{F}=\textbf{f}^1\oplus\textbf{f}^2\oplus ... \oplus\textbf{f}^L
\end{equation}
where $\textbf{F}$ stands for the whole feature vector,
$\textbf{f}^j (j=1,2,...,L)$ is the $j$th layer-wise feature vector, and
$L$ denotes the number of convolutional layers in the applied
base CNN.

Finally, a regression algorithm is applied to map the feature vectors onto
perceptual quality scores. In this study, we made experiments with two
different regression techniques, such as support vector regressor (SVR)
\cite{drucker1997support} 
and Gaussian process regressor (GPR) \cite{williams2006gaussian}.
Specifically, we applied SVR with linear kernel and radial
basis function (RBF) kernel.
GPR was applied with rational quadratic kernel.

%SVR
%with RBF kernel
%it can be expressed as
%\begin{equation}
%y_i = \langle\textbf{w},\varPhi(\textbf{x}_i)\rangle+b,    
%\end{equation}
%where \textbf{w} stands for the separating hyper-plane and $b$ denotes the bias.
%These variables are learned from the training data.
%Moreover, the function $\varPhi(\cdot)$
%maps the original feature space into a higher dimensional one and its inner
%product is a RBF kernel:
%\begin{equation}
%k(\textbf{x}_i, \textbf{x}_j)=\exp\{-\frac{1}{2\sigma^2}
%\lVert\textbf{x}_i - \textbf{x}_j \rVert\},    
%\end{equation}
%where $\sigma$ is a hyper-parameter and corresponds to the scale of the
%RBF kernel; $\textbf{x}_i$
%and $\textbf{x}_j$ are the feature vectors
%belonging to the $i$th and $j$th training images,
%respectively.

\unskip
\unskip
\unskip

\section{Experimental Results}
\label{sec:experimental}
In this section, we present our experimental results and analysis.
First, we introduce the applied evaluation metrics in
Section \ref{sec:metrics}. Second, the implementation details and
the experimental setup are given in Section \ref{sec:setup}.
Subsequently, a detailed parameter study is presented in Section
\ref{sec:parameter}, in which we extensively reason the design
choices of our proposed method.
In Section \ref{sec:distortion}, we explore the performance of our proposed method
on different distortion types and distortion intensity levels.
Subsequently, we examine the relationship between the performance and the
amount of training data in Section \ref{sec:size}.
In Section \ref{sec:comparison},
a comparison to other
state-of-the-art method is carried out using six benchmark IQA
databases, such as
KADID-10k \cite{lin2019kadid}, TID2013 \cite{ponomarenko2015image},
TID2008 \cite{ponomarenko2009tid2008},
\mbox{VCL-FER \cite{zaric2012vcl},} CSIQ~\cite{larson2010most},
and MDID \cite{sun2017mdid}.
The results of the cross database are presented in Section \ref{sec:cross}.
Table~\ref{table:iqadatabase} illustrates some facts about the publicly
available IQA databases used in this paper.~It allows comparisons between the number of reference and test images,
image resolutions, the number of distortion levels, and the number of distortion types.
%In order to prove the generalization ability of the proposed, we
%present a cross-database test in Subsection 
%\ref{sec:cross}.

\begin{table}
\caption{Comparison of publicly available image quality analysis (IQA) databases used in this study.
} %title of the table
\centering % centering table
\begin{tabular}{cccccc}
\toprule
\textbf{Database}&\textbf{Ref. Images}&\textbf{Test Images}&\textbf{Resolution}& \textbf{Distortion Levels}&\textbf{Number of Distortions}\\
    \midrule
TID2008 \cite{ponomarenko2009tid2008}&25&1700&$512\times384$&4&17 \\
CSIQ \cite{larson2010most} & 30& 866&$512\times512$&4--5 &6 \\
VCL-FER \cite{zaric2012vcl} & 23 & 552 &$683\times512$ &6& 4 \\
TID2013 \cite{ponomarenko2015image}&25&3000&$512\times384$&5&24 \\
MDID \cite{sun2017mdid}&20&1600&$512\times384$&4&5 \\
KADID-10k \cite{lin2019kadid} & 81 &10,125&$512\times384$&5&25\\
\bottomrule
\end{tabular}
\label{table:iqadatabase}
\end{table}

\subsection{Evaluation Metrics}
\label{sec:metrics}
A reliable way to evaluate objective FR-IQA methods is based on measuring
the correlation strength between
the ground-truth scores of a publicly available
IQA database and the predicted scores. In the literature,
Pearson's linear correlation
coefficient (PLCC), Spearman's rank-order correlation
coefficient (SROCC), and
Kendall's rank-order correlation coefficient (KROCC) are
widely applied to characterize the
degree of correlation. PLCC between vectors
\textbf{x} and \textbf{y} can be expressed as
\begin{equation}
PLCC(\textbf{x}, \textbf{y}) = \frac{\sum_{i=1}^m (x_i-\overline{x})(y_i-\overline{y})}
{\sqrt{\sum_{i=1}^m (x_i-\overline{x})^2} \sqrt{\sum_{i=1}^m (y_i-\overline{y})^2}}
\end{equation}
where $\overline{x}=\frac{1}{m}\sum_{i=1}^m x_i$
and $\overline{y}=\frac{1}{m}\sum_{i=1}^m y_i$.
Furthermore, \textbf{x} stands for the
vector containing the ground-truth scores,
while \textbf{y} vector consists of the
predicted scores. PLCC performance indices are determined after a non-linear mapping
between objective scores (MOS or differential MOS values) and predicted subjective scores
using a 5-parameter logistic function, as recommended by
the authors of \cite{sheikh2006statistical}.

SROCC between vectors \textbf{x} and \textbf{y} can be defined as
\begin{equation}
SROCC(\textbf{x}, \textbf{y}) = PLCC(\textit{rank(}\textbf{x}\textit{)},
\textit{rank(}\textbf{y}\textit{)})
\end{equation}
where the $rank(\cdot)$ function gives back a vector whose $i$th element
is the rank of the $i$th element in the input vector.
As a consequence, SROCC between vectors \textbf{x} and \textbf{y}
can also be expressed as
\begin{equation}
SROCC(\textbf{x}, \textbf{y}) = \frac{\sum_{i=1}^m (x_i-\hat{x})(y_i-\hat{y})}
{\sqrt{\sum_{i=1}^m (x_i-\hat{x})^2} \sqrt{\sum_{i=1}^m (y_i-\hat{y})^2}}
\end{equation}
where $\hat{x}$ and $\hat{y}$ stand for the middle ranks of 
$\textbf{x}$ and $\textbf{y}$, respectively.

KROCC between vectors \textbf{x} and \textbf{y} can be determined as
\begin{equation}
KROCC(\textbf{x}, \textbf{y}) = \frac{n_c-n_d}{\frac{1}{2}n(n-1)}
\end{equation}
where $n$ is the length of the input vectors, $n_c$ stands for the number of
concordant pairs between $\textbf{x}$
and $\textbf{y}$, and $n_d$ denotes the number of discordant pairs.
\subsection{Experimental Setup}
\label{sec:setup}
KADID-10k \cite{lin2019kadid} was used
to carry out a detailed parameter study
to determine the best design choices of the proposed method.
Subsequently, other publicly available databases, such as
\mbox{TID2013 \cite{ponomarenko2015image},} TID2008 \cite{ponomarenko2009tid2008},
VCL-FER \cite{zaric2012vcl}, CSIQ \cite{larson2010most},
and MDID \cite{sun2017mdid},
were also applied to carry out
a comparison to other state-of-the-art FR-IQA algorithms.
Furthermore, our algorithms and other learning-based state-of-the-art methods were 
evaluated by 5-fold cross-validation with 100
repetitions. 
Specifically,~an~IQA database was divided randomly into a training set (appx. 80\%) and
a test set (appx. 20\%) with respect to the reference, pristine images.
As a consequence, there was no semantic content overlapping between these sets. 
Moreover, we report on the average PLCC,
SROCC, and~KROCC~values. 

All models were implemented and tested in MATLAB R2019a, relying mainly
on the functions of the Deep
Learning Toolbox (formerly Neural Network Toolbox), Statistics and Machine
Learning Toolbox, 
and the Image Processing Toolbox.

\subsection{Parameter Study}
\label{sec:parameter}
\begin{figure}
    \centering
    \begin{subfigure}[b]{0.45\textwidth}
        \includegraphics[width=\textwidth]{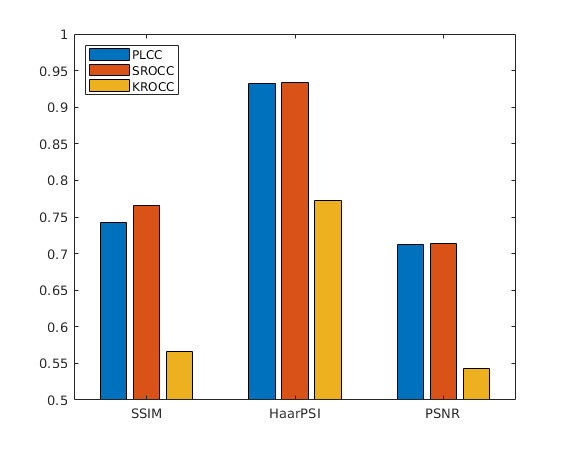}
        \caption{}
    \end{subfigure}
    ~ %add desired spacing between images, e. g. ~, \quad, \qquad, \hfill etc. 
    \begin{subfigure}[b]{0.475\textwidth}
        \includegraphics[width=\textwidth]{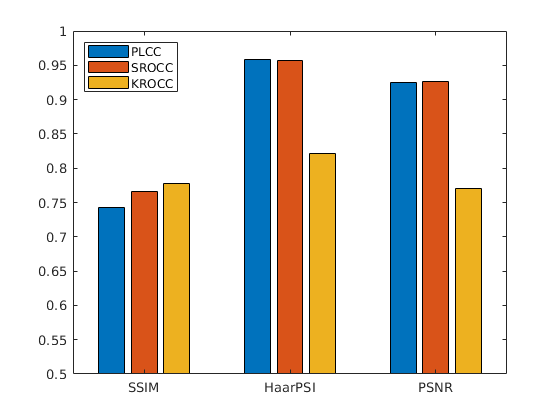}
        \caption{}
    \end{subfigure}

    \quad
    
    \begin{subfigure}[b]{0.45\textwidth}
        \includegraphics[width=\textwidth]{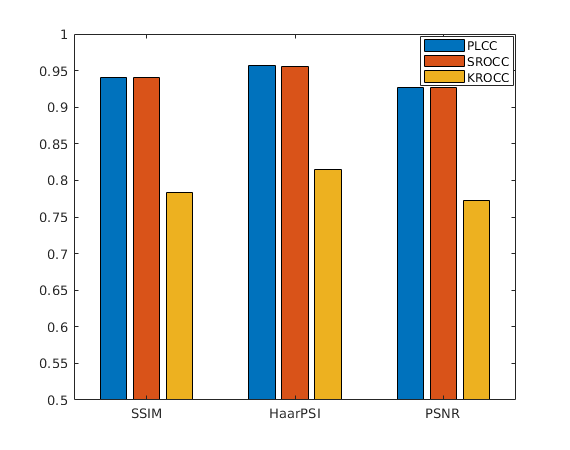}
        \caption{}
    \end{subfigure}
    \caption{Parameter study with respect to the applied regression techniques and image similarity~metrics. (\textbf{a}) Linear SVR. (\textbf{b}) Gaussian SVR. (\textbf{c}) GPR with rational quadratic kernel function.}
    \label{fig:parameterstudy}
\end{figure}

In this subsection, we present a detailed parameter study using the publicly
available KADID-10k~\cite{lin2019kadid} database to find the optimal design
choices of our proposed method. 
Specifically,~we~compared the performance of three traditional metrics
(PSNR, SSIM \cite{wang2004image},
and \mbox{HaarPSI \cite{reisenhofer2018haar}). }
Furthermore, we compared the performance of three different
regression algorithms, such as linear SVR, Gaussian SVR, and GPR
with rational quadratic kernel function.
As already
mentioned, the~evaluation is based on 100 random train-test splits.
Moreover, mean PLCC, SROCC, and KROCC values are reported.
The results of the parameter study are summarized in
Figure \ref{fig:parameterstudy}. From~these results, it can be seen that
HaarPSI metric with Gaussian SVR provides the best results.
This architecture is called \textit{ActMapFeat} in the further
sections.

\subsection{Performance over Different Distortion Types and Levels}
\label{sec:distortion}
In this subsection, we examine the performance of the proposed \textit{ActMapFeat} over different
image distortion types and levels of KADID-10k \cite{lin2019kadid}.~Namely, KADID-10k consists of images with 25 distortion types in 5 levels.
Furthermore, the distortion types can be classified into five groups: blurs, color distortions,
compression, noise, brightness change, spatial distortions, and sharpness and contrast.

The reported mean PLCC, SROCC, and KROCC values were measured
over 100 random train-test splits in Table \ref{table:distortion}.
From these results, it can be observed that \textit{ActMapFeat} is able to perform
relatively uniformly over different image distortion types with the exception of some
color- (color shift, color saturation 1.), brightness- (mean shift),
and patch-related (non-eccentricity patch. color block) noise types. Furthermore, it
performs very well on different blur (Gaussian blur, lens blur, motion blur) and
compression types (JPEG, JPEG2000).

The performance results of \textit{ActMapFeat} over different distortion levels of
KADID-10k \cite{lin2019kadid} are illustrated in Table \ref{table:level}. From these results,
it can be observed that the proposed method performs relatively uniformly over the
different distortion levels. Specifically, it achieves better results on higher
distortion levels than on lower ones. Moreover, the best results can be experienced
at moderate distortion levels.

\begin{table}
\caption{
Mean PLCC, SROCC, KROCC values of the proposed architecture for each
distortion type of KADID-10k \cite{lin2019kadid}. Measured over 100 random train-test splits. The distortion
types found in KADID-10k can be classified into five groups: blurs, color distortions, compression, noise, brightness change, spatial~distortions, and
sharpness and contrast.
} %title of the table
\centering % centering table
\begin{center}
  \begin{tabular}{c c c c}
\toprule
% & \multicolumn{2}{c | }{KoNViD-1k \cite{343729:7601552}} & \multicolumn{2}{c|}{LIVE VQA \cite{343729:7652544}} \\
%\cline{2-5}
\textbf{Distortion} & \textbf{PLCC}    & \textbf{SROCC}  & \textbf{KROCC}\\
\midrule
 Gaussian blur                                & 0.987  & 0.956  & 0.828  \\
 Lens blur                                    & 0.971  & 0.923  & 0.780  \\
 Motion blur                                  & 0.976  & 0.960  & 0.827  \\
\midrule
 Color diffusion                              & 0.971  & 0.906  & 0.744  \\
 Color shift                                  & 0.942  & 0.866  & 0.698  \\
 Color quantization                           & 0.902  & 0.868  & 0.692  \\
 Color saturation 1.                          & 0.712  & 0.654  & 0.484  \\
 Color saturation 2.                          & 0.973  & 0.945  & 0.798 \\
\midrule
 JPEG2000                                     & 0.977 & 0.941  & 0.800  \\
 JPEG                                         & 0.983 & 0.897  & 0.741  \\
\midrule
 White noise                                  & 0.921 & 0.919  & 0.758  \\
 White noise in color component               & 0.958 & 0.946  & 0.802  \\
 Impulse noise                                & 0.875 & 0.872  & 0.694  \\
 Multiplicative noise                         & 0.958 & 0.952  & 0.813 \\
 Denoise                                      & 0.955 & 0.941  & 0.799 \\
\midrule
 Brighten                                     & 0.969  & 0.951  & 0.815\\
 Darken                                       & 0.973  & 0.919  & 0.769  \\
 Mean shift                                   & 0.778  & 0.777  & 0.586  \\
\midrule
 Jitter                                       & 0.981  & 0.962  & 0.834  \\
 Non-eccentricity patch                       & 0.693  & 0.667  & 0.489  \\
 Pixelate                                     & 0.909  & 0.854  & 0.681  \\
 Quantization                                 & 0.893  & 0.881  & 0.705  \\
 Color block                                  & 0.647  & 0.539  & 0.386  \\
\midrule
 High sharpen                                 & 0.948  & 0.938  & 0.786  \\
 Contrast change                              & 0.802  & 0.805  & 0.607  \\
\midrule
 \textbf{All}                                 & 0.959  & 0.957  & 0.819  \\
\bottomrule
\end{tabular}  
\end{center}
\label{table:distortion}
\end{table}
\unskip
\begin{table}
\caption{
Mean PLCC, SROCC, KROCC values of the proposed  architecture for each distortion level of
KADID-10k \cite{lin2019kadid}. Measured over 100 random train-test splits. KADID-10k \cite{lin2019kadid}
contains images with five different distortion levels, where Level $1$ stands for the lowest
amount of distortion, while Level $5$ denotes the highest amount.
} %title of the table
\centering % centering table
\begin{center}
  \begin{tabular}{c c cc}
\toprule
% & \multicolumn{2}{c | }{KoNViD-1k \cite{343729:7601552}} & \multicolumn{2}{c|}{LIVE VQA \cite{343729:7652544}} \\
%\cline{2-5}
 \textbf{Level of Distortion} & \textbf{PLCC} & \textbf{SROCC} & \textbf{KROCC}\\
\midrule
 Level $1$  & 0.889  & 0.843  & 0.659  \\
 Level $2$  & 0.924  & 0.918  & 0.748  \\
 Level $3$  & 0.935  & 0.933  & 0.777  \\
 Level $4$  & 0.937  & 0.922  & 0.765  \\
 Level $5$  & 0.931  & 0.897  & 0.725  \\
\midrule
 \textbf{All}& 0.959 & 0.957  & 0.819  \\
\bottomrule
\end{tabular}  
\end{center}
\label{table:level}
\end{table}

%\begin{figure}[H]
%\centering
%\includegraphics[width=15cm]{TrainRatio.png}
%\caption{Plot of mean PLCC, SROCC, and KROCC between ground-truth and predicted
%MOS values measured on KADID-10k \cite{lin2019kadid} over 100 random train-test splits
%as a function of the percentage of the dataset used for training.}
%\label{fig:percentage}
%\end{figure} 

%\begin{figure}[t!]
%    \centering
%    \begin{subfigure}{0.425\textwidth}
%        \centering
%        \includegraphics[height=1.95in]{PLCC_Dataset.png}
%        \caption{Mean PLCC as a function of the percentage of the dataset
%        used for training.}
%    \end{subfigure}%
%    ~ 
%    \begin{subfigure}{0.425\textwidth}
%        \centering
%        \includegraphics[height=1.95in]{SROCC_Dataset.png}
%        \caption{Mean SROCC as a function of the percentage of the dataset
%        used for training.}
%    \end{subfigure}
    
%    \centering
%    \begin{subfigure}{0.425\textwidth}
%        \centering
%        \includegraphics[height=1.95in]{KROCC_Dataset.png}
%        \caption{Mean KROCC as a function of the percentage of the dataset
%        used for training.}
%    \end{subfigure}%
%    \caption{Plot of mean PLCC (a), SROCC (b), and KROCC (c) between
%    ground-truth and predicted MOS values measured on KADID-10k
%    over 20 random train-test splits as a function of the percentage
%    of the dataset used for training.}
%    \label{fig:percentage}
%\end{figure}

\subsection{Effect of the Training Set Size}
\label{sec:size}
\begin{figure}
    \centering
    \begin{subfigure}[b]{0.45\textwidth}
        \includegraphics[width=\textwidth]{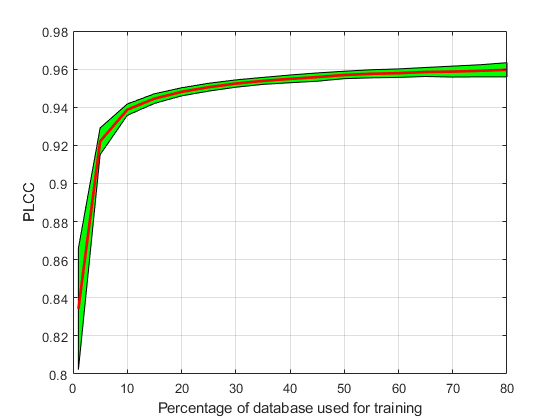}
        \caption{}
    \end{subfigure}
    ~ %add desired spacing between images, e. g. ~, \quad, \qquad, \hfill etc. 
    \begin{subfigure}[b]{0.45\textwidth}
        \includegraphics[width=\textwidth]{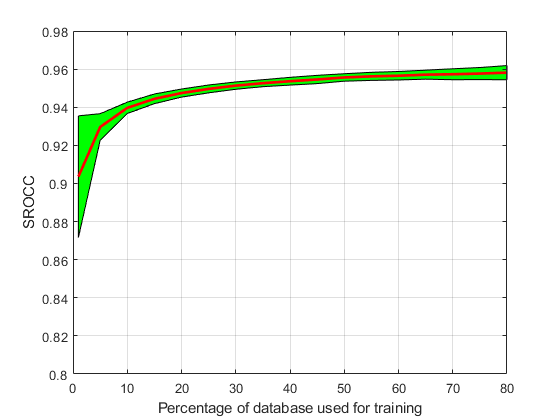}
        \caption{}
    \end{subfigure}

    \quad
    
    \begin{subfigure}[b]{0.45\textwidth}
        \includegraphics[width=\textwidth]{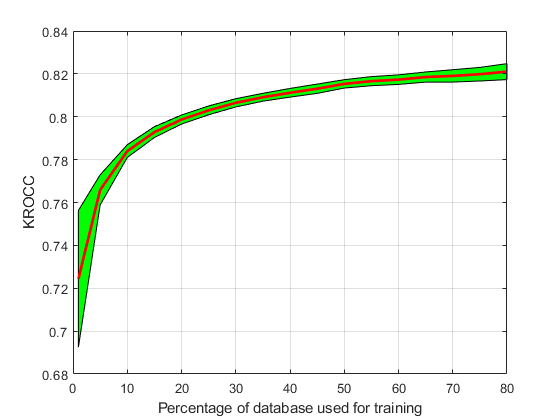}
        \caption{}
    \end{subfigure}
    \caption{Plots of mean PLCC, SROCC, and KROCC between ground-truth and predicted
MOS values measured on KADID-10k \cite{lin2019kadid} over 100 random train-test splits
as a function of the percentage of the dataset used for training. Mean PLCC,
SROCC, and KROCC are plotted as red lines, while the standard deviations are depicted
as shaded areas. (\textbf{a}) Mean PLCC as a function of the dataset used for training. (\textbf{b}) Mean SROCC as a function of the dataset used for training. (\textbf{c}) Mean KROCC as a function of the dataset used for training.}
    \label{fig:percentage}
\end{figure}

In general, the number of training images has a strong impact on the
performance
of machine/deep
learning systems \cite{lin2018koniq,lin2019kadid,cho2015much}.
In this subsection,
we study the relationship between the
number of training images and the performance
using the KADID-10k \cite{lin2019kadid} database.
In our experiments, the ratio of the training images in the database varied
from 5\% to 80\%, while at the same time, those of the test images varied from 95\% to 20\%.
The results are illustrated in Figure \ref{fig:percentage}. It can be observed that the
proposed system is rather robust to the size of the training set. 
Specifically,~the~mean PLCC, SROCC, and KROCC are $0.923$, $0.929$, and $0.765$, if the ratio of
the training images is 5\%. These performance metrics increase to $0.959$, $0.957$, and
$0.821$, respectively, when the ratio of the training set reaches 80\%, which is a common choice in
machine learning. On the whole, our system can be trained with few amount of data to reach
relatively high PLCC, SROCC, and KROCC values. This~proves the effectiveness of the proposed
feature extraction method from distorted reference image~pairs.

\subsection{Comparison to the State-of-the-Art}
\label{sec:comparison}
Our proposed algorithm was compared to several state-of-the-art
FR-IQA metrics, including
SSIM~\cite{wang2004image}, MS-SSIM \cite{wang2003multiscale},
MAD \cite{larson2010most}, GSM \cite{liu2011image},
HaarPSI \cite{reisenhofer2018haar},
MDSI \cite{nafchi2016mean}, CSV \cite{temel2016csv},
GMSD \cite{xue2013gradient}, DSS~\cite{balanov2015image}, VSI~\cite{zhang2014vsi},
PerSIM \cite{temel2015persim}, BLeSS-SR-SIM \cite{temel2016bless},
BLeSS-FSIM \cite{temel2016bless}, BLeSS-FSIMc \cite{temel2016bless},
LCSIM1 \cite{oszust2016full}, ReSIFT~\cite{temel2016resift}, IQ($\mathcal{L}_T$) \cite{ali2017image},
MS-UNIQUE \cite{prabhushankar2017ms}, RVSIM \cite{yang2018rvsim},
2stepQA \cite{yu2019predicting},
SUMMER \cite{temel2019perceptual},
CEQI \cite{layek2019center}, CEQIc \cite{layek2019center},
VCGS \cite{shi2020full}, and DISTS \cite{ding2020image},
whose original source code are available
online.
Moreover, we reimplemented SSIM CNN \cite{amirshahi2018reviving} in
MATLAB R2019a \protect ({Available
: \url{https://github.com/Skythianos/SSIM-CNN}}).
For learning-based approaches, we retrained the models using exactly the same database
partition (approx. 80\% for training and 20\% for testing with respect to the
reference images to avoid semantic overlap) that
we used for our method. 
Since the feature extraction part of LCSIM1 \cite{oszust2016full} was not given,
we can report the results of LCSIM1 on
TID2013 \cite{ponomarenko2015image}, CSIQ \cite{larson2010most},
and TID2008 \cite{ponomarenko2009tid2008} databases.
Furthermore, mean PLCC, SROCC, and KROCC values were
measured over 100 random train-test splits for machine learning-based algorithms.
In contrast, traditional FR-IQA metrics are
tested on the whole database, and we report on the PLCC, SROCC, and KROCC values.
Besides FR-IQA methods, some recently published deep learning-based NR-IQA algorithms, including \mbox{DeepFL-IQA \cite{lin2020deepfl},}
BLINDER \cite{gao2018blind}, RankIQA \cite{liu2017rankiqa},
BPSOM-MD \cite{pan2018blind}, and NSSADNN \cite{yan2019naturalness},
have been added to our comparison. Due to the difficulty of reimplementation of
deep NR-IQA methods, the performance numbers in the corresponding
papers are reported in this~study.

The results of the performance comparison to the state-of-the-art on
KADID-10k \cite{lin2019kadid}, TID2013~\cite{ponomarenko2015image},
VCL-FER \cite{zaric2012vcl},
TID2008 \cite{ponomarenko2009tid2008},
MDID \cite{sun2017mdid}, and CSIQ \cite{larson2010most} are summarized in
Tables \ref{table:iqacomp1}--\ref{table:iqacomp4}, respectively. It~can be observed that
the performance of
the examined state-of-the-art FR-IQA algorithms are
far from perfect on KADID-10k \cite{lin2019kadid}.
In contrast, our method was able to produce PLCC and SROCC values over 0.95.
Furthermore, our KROCC
value is about 0.09 higher than the second the best one. On~the smaller
TID2013 \cite{ponomarenko2015image}, TID2008 \cite{ponomarenko2009tid2008},
MDID \cite{sun2017mdid}, and VCL-FER \cite{zaric2012vcl} 
IQA databases, the
performances of the examined state-of-the-art
approaches significantly improve. In spite of this, our
method also gives the best results on these databases in terms of
PLCC, SROCC, and KROCC, as well.
Similarly, our method achieves the best results on CSIQ \cite{larson2010most}.
On the other hand, the difference
between the proposed method and other state-of-the-art methods is
observably less than those on the other IQA benchmark databases.
Significance tests were also carried out to prove that the achieved improvements on
benchmark data sets are
significant.
More precisely, the
ITU (International Telecommunication Union)
guidelines \cite{itu20121401} for
evaluating quality models were
followed. The $H_0$ hypothesis for a given correlation
coefficient (PLCC, SROCC, or KROCC) was that a rival state-of-the-art
method produces not significantly different values with $p<0.05$. Moreover, the
variances of the $z$-transforms were determined as $1.06/(N-3)$, where
$N$ stands for the number of images in a given IQA database.
In Tables \ref{table:iqacomp1}--\ref{table:iqacomp4},
the \colorbox{green!30}{green} background color stands for
the fact that the correlation is lower than
those of the proposed method and the difference is statistically.

\begin{figure}
    \centering
    \begin{subfigure}[b]{0.45\textwidth}
        \includegraphics[width=\textwidth]{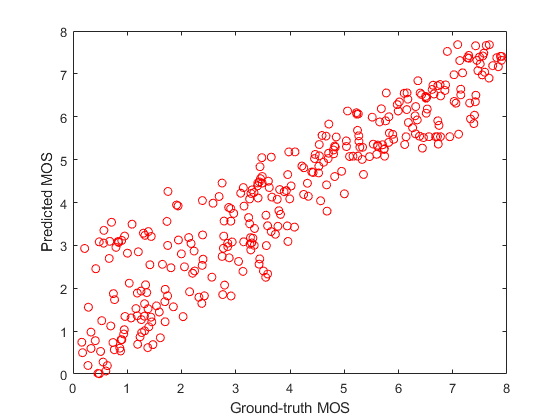}
        \caption{}
    \end{subfigure}
    ~ %add desired spacing between images, e. g. ~, \quad, \qquad, \hfill etc. 
    \begin{subfigure}[b]{0.45\textwidth}
        \includegraphics[width=\textwidth]{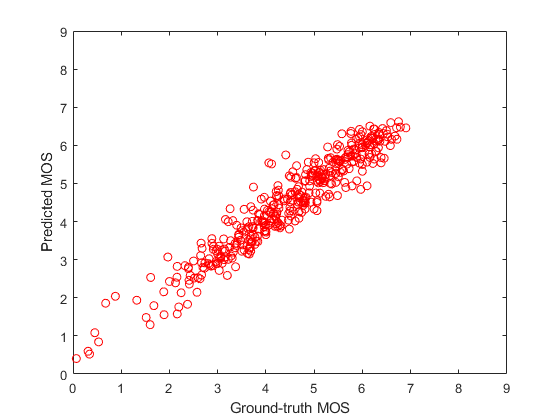}
        \caption{}
    \end{subfigure}

    \quad
    
    \begin{subfigure}[b]{0.45\textwidth}
        \includegraphics[width=\textwidth]{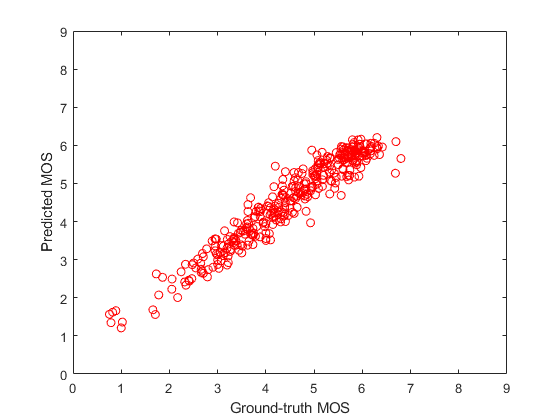}
        \caption{}
    \end{subfigure}
    ~
    \begin{subfigure}[b]{0.45\textwidth}
        \includegraphics[width=\textwidth]{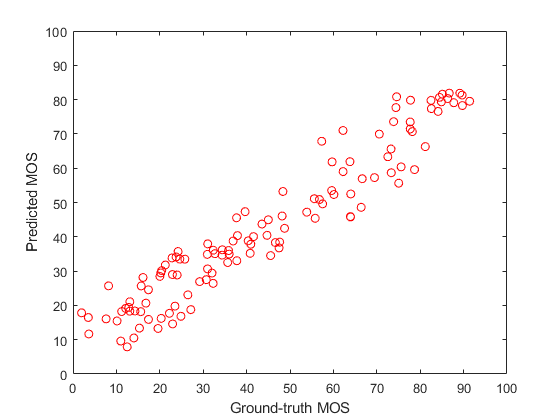}
        \caption{}
    \end{subfigure}

    \quad

    \unskip
    \begin{subfigure}[b]{0.45\textwidth}
        \includegraphics[width=\textwidth]{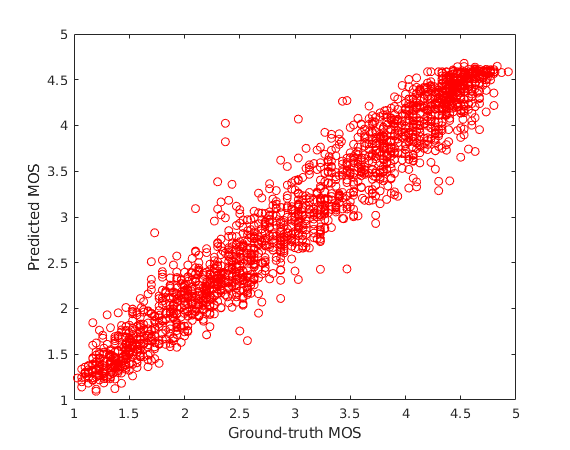}
        \caption{}
    \end{subfigure}
    ~
    \begin{subfigure}[b]{0.45\textwidth}
        \includegraphics[width=\textwidth]{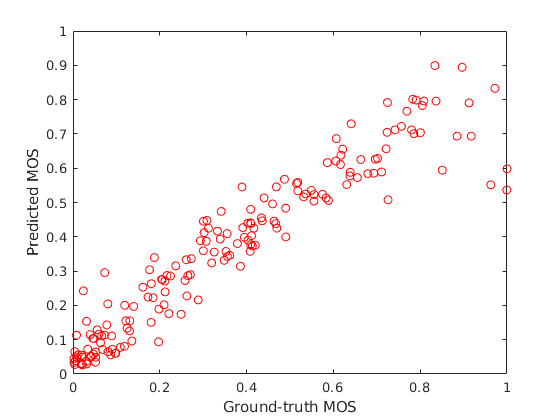}
        \caption{}
    \end{subfigure}
    \caption{Scatter plots of the ground-truth MOS against the predicted MOS of \textit{ActMapFeat} on different test sets of MDID \cite{sun2017mdid}, TID2008 \cite{ponomarenko2009tid2008}, TID2013 \cite{ponomarenko2015image}, VCL-FER \cite{zaric2012vcl}, KADID-10k \cite{lin2019kadid}, and
    CSIQ \cite{larson2010most} IQA benchmark databases. (\textbf{a}) MDID \cite{sun2017mdid}. (\textbf{b}) TID2008 \cite{ponomarenko2009tid2008}. (\textbf{c}) TID2013 \cite{ponomarenko2015image}. (\textbf{d}) VCL-FER \cite{zaric2012vcl}. (\textbf{e}) KADID-10k \cite{lin2019kadid}. (\textbf{f}) CSIQ \cite{larson2010most}.}
    \label{fig:scatter}
\end{figure}

\begin{figure}
    \centering
    \begin{subfigure}[b]{0.45\textwidth}
        \includegraphics[width=\textwidth]{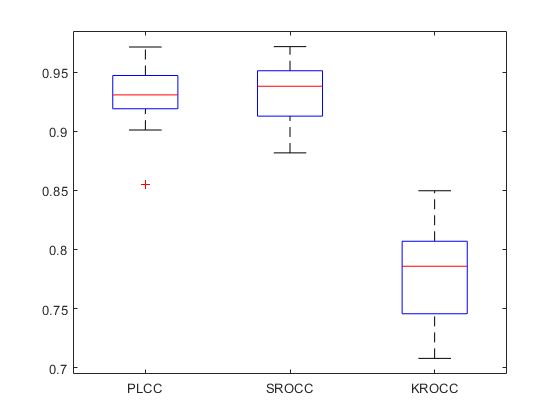}
        \caption{}
    \end{subfigure}
    ~ %add desired spacing between images, e. g. ~, \quad, \qquad, \hfill etc. 
    \begin{subfigure}[b]{0.45\textwidth}
        \includegraphics[width=\textwidth]{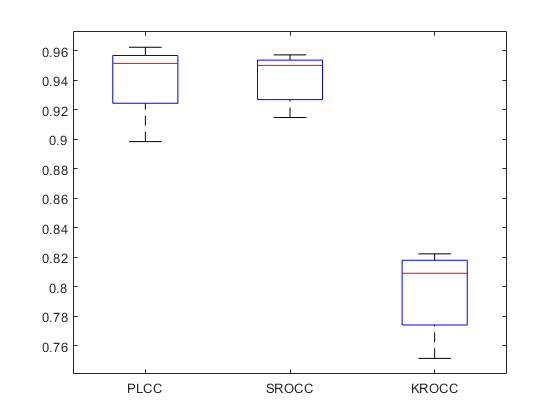}
        \caption{}
    \end{subfigure}

    \quad
    
    \begin{subfigure}[b]{0.45\textwidth}
        \includegraphics[width=\textwidth]{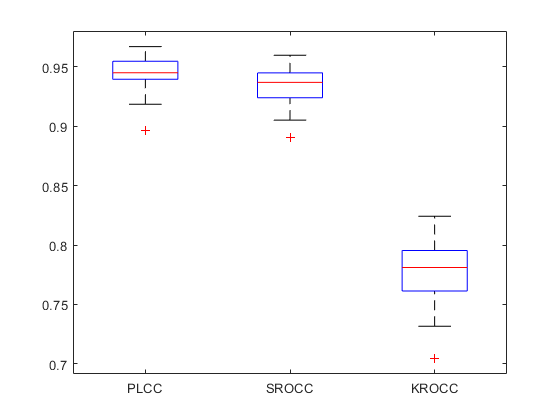}
        \caption{}
    \end{subfigure}
    ~
    \begin{subfigure}[b]{0.45\textwidth}
        \includegraphics[width=\textwidth]{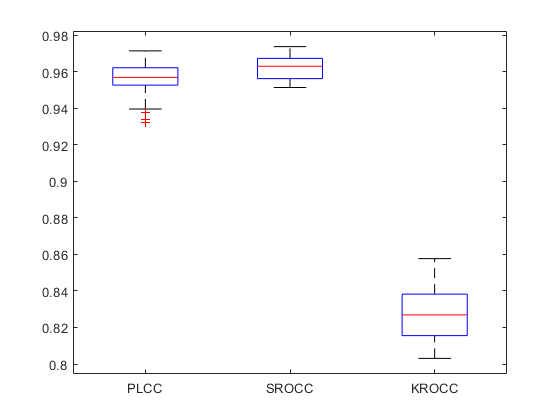}
        \caption{}
    \end{subfigure}

    \quad

    \begin{subfigure}[b]{0.45\textwidth}
        \includegraphics[width=\textwidth]{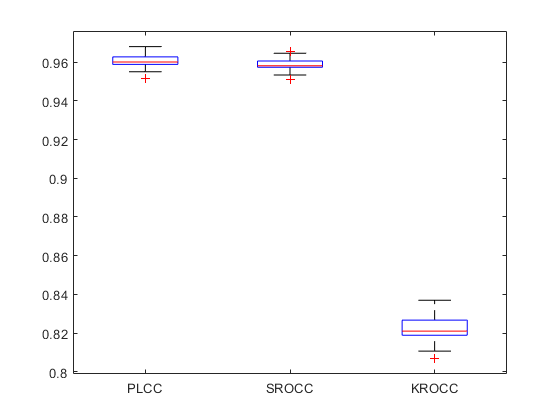}
        \caption{}
    \end{subfigure}
    ~
    \begin{subfigure}[b]{0.45\textwidth}
        \includegraphics[width=\textwidth]{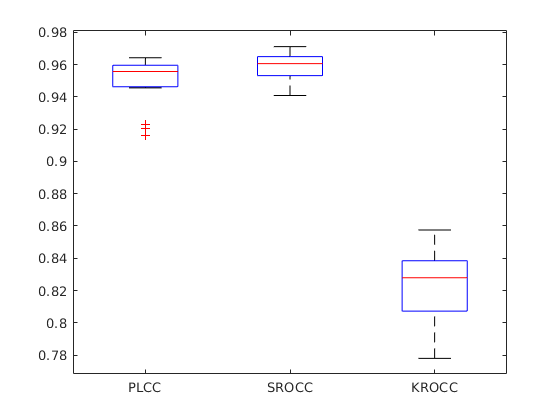}
        \caption{}
    \end{subfigure}
    \caption{Box plots of the measured PLCC, SROCC, and KROCC values produced by
    the proposed \textit{ActMapFeat}
    over 100 random train-test splits on MDID \cite{sun2017mdid}, TID2008 \cite{ponomarenko2009tid2008},
   \mbox{ TID2013 \cite{ponomarenko2015image},} \mbox{VCL-FER \cite{zaric2012vcl}, }KADID-10k \cite{lin2019kadid}, and
    CSIQ \cite{larson2010most} IQA benchmark databases. (\textbf{a}) MDID \cite{sun2017mdid}. \mbox{(\textbf{b}) TID2008 \cite{ponomarenko2009tid2008}.} (\textbf{c}) TID2013 \cite{ponomarenko2015image}. (\textbf{d}) VCL-FER \cite{zaric2012vcl}. (\textbf{e}) KADID-10k \cite{lin2019kadid}. (\textbf{f}) CSIQ \cite{larson2010most}.}
    \label{fig:box}
\end{figure}

Figure \ref{fig:scatter} depicts scatter plots
of ground-truth MOS values against predicted
MOS values on MDID~\cite{sun2017mdid}, TID2008 \cite{ponomarenko2009tid2008},
TID2013 \cite{ponomarenko2015image}, VCL-FER \cite{zaric2012vcl},
KADID-10k \cite{lin2019kadid}, and CSIQ \cite{larson2010most} test sets.
Figure~\ref{fig:box} depicts the box plots of the measured PLCC, SROCC, and
KROCC values over 100 random train-test splits. On each box, the central mark
denotes the median, and the bottom and top edges of the box represent the $25$th
and $75$th percentiles, respectively. Moreover, the whiskers extend to the most
extreme values which are not considered outliers.

%Figure \ref{fig:Scatter} depicts scatter
%plots showing the ground-truth MOS values against predicted MOS
%values on KADID-10k \cite{lin2019kadid},
%TID2013 \cite{ponomarenko2015image},
%TID2008 \cite{ponomarenko2009tid2008},
%and MDID \cite{sun2017mdid} test sets,
%respectively. From these scatter plots, it can be seen intuitively which was carefully
%measured in Table \ref{table:level}. Namely, our method is able to perform uniformly
%over different distortion levels.

\begin{table}
\caption{Performance comparison on KADID-10k \cite{lin2019kadid}
and TID2013 \cite{ponomarenko2015image}
databases.
Mean PLCC, SROCC, and KROCC values
are reported for the learning-based approaches
measured over 100 random train-test splits.
The best results are typed in \textbf{bold}.
The
\colorbox{green!30}{green} background color stands for the fact that the correlation is
lower than those of the proposed method and the difference is statistically
significant with \emph{p} < 0.05. We used '-' if the data were not available.
} %title of the table
\centering % centering table
\begin{center}
  \begin{tabular}{c  c  c  c c  c  c}
\toprule
 & \multicolumn{3}{c  }{\textbf{ KADID-10k \cite{lin2019kadid}}}
 & \multicolumn{3}{c  }{\textbf{ TID2013 \cite{ponomarenko2015image}}}\\%  & 
% \multicolumn{3}{c}{\textbf{ TID2013 \cite{ponomarenko2015image}}} \\
\cline{2-7}
 &  \textbf{PLCC} &  \textbf{SROCC}  &  \textbf{KROCC}
 &  \textbf{PLCC} &  \textbf{SROCC}  &  \textbf{KROCC}\\
\hline
% PSNR} &  0.653} &  0.653} &  0.518} & 0.647} & 0.647} & 0.488} \\
SSIM \cite{wang2004image}&\colorbox{green!30}{0.670}&\colorbox{green!30}{0.671}& \colorbox{green!30}{0.489}&\colorbox{green!30}{0.618}&\colorbox{green!30}{0.616}&\colorbox{green!30}{0.437}\\
MS-SSIM \cite{wang2003multiscale} &\colorbox{green!30}{0.819} &\colorbox{green!30}{0.821} &\colorbox{green!30}{0.630}&\colorbox{green!30}{0.794}&\colorbox{green!30}{0.785}&\colorbox{green!30}{0.604} \\
MAD \cite{larson2010most}&\colorbox{green!30}{0.716} &\colorbox{green!30}{0.724} &\colorbox{green!30}{0.535}&\colorbox{green!30}{0.827}&\colorbox{green!30}{0.778}&\colorbox{green!30}{0.600} \\
GSM \cite{liu2011image}&\colorbox{green!30}{0.780} &\colorbox{green!30}{0.780} &\colorbox{green!30}{0.588}&\colorbox{green!30}{0.789}&\colorbox{green!30}{0.787}&\colorbox{green!30}{0.593} \\
HaarPSI \cite{reisenhofer2018haar} & \colorbox{green!30}{0.871} & \colorbox{green!30}{0.885} & \colorbox{green!30}{0.699}&\colorbox{green!30}{0.886}&\colorbox{green!30}{0.859}&\colorbox{green!30}{0.678}\\
MDSI \cite{nafchi2016mean}   & \colorbox{green!30}{0.887} & \colorbox{green!30}{0.885} & \colorbox{green!30}{0.702}&\colorbox{green!30}{0.867}&\colorbox{green!30}{0.859}&\colorbox{green!30}{0.677}\\
CSV \cite{temel2016csv}&\colorbox{green!30}{0.671}&\colorbox{green!30}{0.669}& \colorbox{green!30}{0.531}&\colorbox{green!30}{0.852}&\colorbox{green!30}{0.848}&\colorbox{green!30}{0.657}  \\
GMSD \cite{xue2013gradient}&\colorbox{green!30}{0.847}&\colorbox{green!30}{0.847} & \colorbox{green!30}{0.664}&\colorbox{green!30}{0.846}&\colorbox{green!30}{0.844}&\colorbox{green!30}{0.663}  \\
DSS \cite{balanov2015image}&\colorbox{green!30}{0.855}&\colorbox{green!30}{0.860} & \colorbox{green!30}{0.674}&\colorbox{green!30}{0.793}&\colorbox{green!30}{0.781}&\colorbox{green!30}{0.604}\\
VSI \cite{zhang2014vsi}&\colorbox{green!30}{0.874} & \colorbox{green!30}{0.861} & \colorbox{green!30}{0.678}&\colorbox{green!30}{0.900}&\colorbox{green!30}{0.894}&\colorbox{green!30}{0.677} \\
PerSIM \cite{temel2015persim} & \colorbox{green!30}{0.819} & \colorbox{green!30}{0.824} & \colorbox{green!30}{0.634}&\colorbox{green!30}{0.825}&\colorbox{green!30}{0.826}&\colorbox{green!30}{0.655}\\
BLeSS-SR-SIM \cite{temel2016bless}& \colorbox{green!30}{0.820} & \colorbox{green!30}{0.824} & \colorbox{green!30}{0.633}&\colorbox{green!30}{0.814}&\colorbox{green!30}{0.828}&\colorbox{green!30}{0.648} \\
BLeSS-FSIM \cite{temel2016bless}& \colorbox{green!30}{0.814} & \colorbox{green!30}{0.816} & \colorbox{green!30}{0.624}&\colorbox{green!30}{0.824}&\colorbox{green!30}{0.830}&\colorbox{green!30}{0.649} \\
BLeSS-FSIMc \cite{temel2016bless}& \colorbox{green!30}{0.845} & \colorbox{green!30}{0.848} & \colorbox{green!30}{0.658}&\colorbox{green!30}{0.846}&\colorbox{green!30}{0.849}&\colorbox{green!30}{0.667} \\
LCSIM1 \cite{oszust2016full} & - & - & - &\colorbox{green!30}{0.914}&\colorbox{green!30}{0.904}&\colorbox{green!30}{0.733}\\
ReSIFT \cite{temel2016resift} &\colorbox{green!30}{0.648} &\colorbox{green!30}{0.628} &\colorbox{green!30}{0.468} &\colorbox{green!30}{0.630}&\colorbox{green!30}{0.623}&\colorbox{green!30}{0.471} \\
IQ($\mathcal{L}_T$) \cite{ali2017image} & \colorbox{green!30}{0.853} & \colorbox{green!30}{0.852} & \colorbox{green!30}{0.641} &\colorbox{green!30}{0.844}&\colorbox{green!30}{0.842}&\colorbox{green!30}{0.631}\\
MS-UNIQUE \cite{prabhushankar2017ms} & \colorbox{green!30}{0.845} & \colorbox{green!30}{0.840} & \colorbox{green!30}{0.648}&\colorbox{green!30}{0.865}&\colorbox{green!30}{0.871}&\colorbox{green!30}{0.687} \\
SSIM CNN \cite{amirshahi2018reviving}& \colorbox{green!30}{0.811}& \colorbox{green!30}{0.814}& \colorbox{green!30}{0.630}&\colorbox{green!30}{0.759}&\colorbox{green!30}{0.752}&\colorbox{green!30}{0.566}\\
RVSIM \cite{yang2018rvsim}&\colorbox{green!30}{0.728}&\colorbox{green!30}{0.719}&\colorbox{green!30}{0.540}&\colorbox{green!30}{0.763}&\colorbox{green!30}{0.683}&\colorbox{green!30}{0.520}\\
2stepQA \cite{yu2019predicting}& \colorbox{green!30}{0.768}& \colorbox{green!30}{0.771}& \colorbox{green!30}{0.571}&\colorbox{green!30}{0.736}&\colorbox{green!30}{0.733}&\colorbox{green!30}{0.550}\\
SUMMER \cite{temel2019perceptual} & \colorbox{green!30}{0.719} & \colorbox{green!30}{0.723} & \colorbox{green!30}{0.540}&\colorbox{green!30}{0.623} &\colorbox{green!30}{0.622} & \colorbox{green!30}{0.472} \\
CEQI \cite{layek2019center}  & \colorbox{green!30}{0.862}&\colorbox{green!30}{0.863} &\colorbox{green!30}{0.681} &\colorbox{green!30}{0.855}&\colorbox{green!30}{0.802} &\colorbox{green!30}{0.635} \\
CEQIc \cite{layek2019center} &\colorbox{green!30}{0.867} &\colorbox{green!30}{0.864} &\colorbox{green!30}{0.682} &\colorbox{green!30}{0.858} &\colorbox{green!30}{0.851} &\colorbox{green!30}{0.638} \\
VCGS \cite{shi2020full}& \colorbox{green!30}{0.873} & \colorbox{green!30}{0.871} & \colorbox{green!30}{0.683}&\colorbox{green!30}{0.900} &\colorbox{green!30}{0.893} & \colorbox{green!30}{0.712}\\
DISTS \cite{ding2020image} & \colorbox{green!30}{0.809}&\colorbox{green!30}{0.814} &\colorbox{green!30}{0.626} &\colorbox{green!30}{0.759} &\colorbox{green!30}{0.711} &\colorbox{green!30}{0.524} \\
%SSIM CNN \cite{amirshahi2018reviving}& 0.646 & 0.848 & 0.699 \\
%DeepQA \cite{kim2017deep}& 0.891/0.891 $(\pm 0.008)$ & 0.897/0.897 $(\pm 0.009)$ &0.734/0.735 $(\pm 0.009)$ \\
%WaDIQaM-FR \cite{bosse2017deep}& 0.889/0.890 $(\pm 0.008)$ & 0.896/0.897 $(\pm 0.008)$ &0.736/0.737 $(\pm 0.009)$ \\
\hline
DeepFL-IQA \cite{lin2020deepfl}&0.938 &0.936 &- &0.876 &0.858 &- \\
BLINDER \cite{gao2018blind}   &- &- &- &0.819 &0.838 &- \\
RankIQA \cite{liu2017rankiqa}   &- &- &- &0.799 &0.780 &- \\
BPSOM-MD \cite{pan2018blind}  &- &- &- &0.879 &0.863 &- \\
NSSADNN \cite{yan2019naturalness}   &- &- &- &0.910 &0.844 &- \\
\midrule
 \textit{ ActMapFeat} (ours) & \textbf{0.959} & \textbf{0.957}  & \textbf{0.819}
 &\textbf{0.943} &\textbf{0.936} &\textbf{0.780}\\
\bottomrule
\end{tabular}  
\end{center}
\label{table:iqacomp1}
\end{table}
\unskip
\begin{table}
\caption{Performance comparison on VCL-FER \cite{zaric2012vcl} and
TID2008 \cite{ponomarenko2009tid2008}
databases.
Mean PLCC, SROCC, and KROCC values
are reported for the learning-based approaches
measured over 100 random train-test splits.
The best results are typed in \textbf{bold}.
The
\colorbox{green!30}{green} background color stands for the fact that the correlation is
lower than those of the proposed method and the difference is statistically
significant with \emph{p} < 0.05. We used '-' if the data were not available.} %title of the table
\centering % centering table
\begin{center}
  \begin{tabular}{c  c  c  cccc}
\toprule
 & \multicolumn{3}{c  }{\textbf{ VCL-FER \cite{zaric2012vcl}}}
 & \multicolumn{3}{c  }{\textbf{ TID2008 \cite{ponomarenko2009tid2008}}}\\%  & 
% \multicolumn{3}{c}{\textbf{ TID2013 \cite{ponomarenko2015image}}} \\
\cline{2-7}
 &  \textbf{PLCC} &  \textbf{SROCC}  &  \textbf{KROCC}
 &  \textbf{PLCC} &  \textbf{SROCC}  &  \textbf{KROCC}\\
\hline
% PSNR} &  0.653} &  0.653} &  0.518} & 0.647} & 0.647} & 0.488} \\
SSIM \cite{wang2004image}&\colorbox{green!30}{0.751}&\colorbox{green!30}{0.859}&\colorbox{green!30}{0.666}&\colorbox{green!30}{0.669}&\colorbox{green!30}{0.675}&\colorbox{green!30}{0.485}\\
MS-SSIM \cite{wang2003multiscale} &\colorbox{green!30}{0.917} &\colorbox{green!30}{0.925} &\colorbox{green!30}{0.753} &\colorbox{green!30}{0.838} &\colorbox{green!30}{0.846} &\colorbox{green!30}{0.648} \\
MAD \cite{larson2010most}&\colorbox{green!30}{0.904} &\colorbox{green!30}{0.906} &\colorbox{green!30}{0.721}&\colorbox{green!30}{0.831}&\colorbox{green!30}{0.829}&\colorbox{green!30}{0.639} \\
GSM \cite{liu2011image}&\colorbox{green!30}{0.904} &\colorbox{green!30}{0.905} &\colorbox{green!30}{0.721}&\colorbox{green!30}{0.782}&\colorbox{green!30}{0.781}&\colorbox{green!30}{0.578} \\
HaarPSI \cite{reisenhofer2018haar} &\colorbox{green!30}{0.938}&\colorbox{green!30}{0.946}&\colorbox{green!30}{0.789}&\colorbox{green!30}{0.916}&\colorbox{green!30}{0.897}&\colorbox{green!30}{0.723}  \\
MDSI \cite{nafchi2016mean} &\colorbox{green!30}{0.935}&\colorbox{green!30}{0.939}&\colorbox{green!30}{0.774}&\colorbox{green!30}{0.877}&\colorbox{green!30}{0.892}&\colorbox{green!30}{0.724}\\
CSV \cite{temel2016csv}&\colorbox{green!30}{0.951}&\colorbox{green!30}{0.952}&\colorbox{green!30}{0.798}&\colorbox{green!30}{0.852}&\colorbox{green!30}{0.851}&\colorbox{green!30}{0.659}  \\
GMSD \cite{xue2013gradient}&\colorbox{green!30}{0.918}&\colorbox{green!30}{0.918}&\colorbox{green!30}{0.741}&\colorbox{green!30}{0.879}&\colorbox{green!30}{0.879}&\colorbox{green!30}{0.696}\\
DSS \cite{balanov2015image}&\colorbox{green!30}{0.925}&\colorbox{green!30}{0.927}&\colorbox{green!30}{0.757}&\colorbox{green!30}{0.860}&\colorbox{green!30}{0.860}&\colorbox{green!30}{0.672}\\
VSI \cite{zhang2014vsi}&\colorbox{green!30}{0.929}&\colorbox{green!30}{0.932}&\colorbox{green!30}{0.763}&\colorbox{green!30}{0.898}&\colorbox{green!30}{0.896}&\colorbox{green!30}{0.709}\\
PerSIM \cite{temel2015persim}&\colorbox{green!30}{0.926}&\colorbox{green!30}{0.928}&\colorbox{green!30}{0.761}&\colorbox{green!30}{0.826}&\colorbox{green!30}{0.830}&\colorbox{green!30}{0.655}\\
BLeSS-SR-SIM \cite{temel2016bless}&\colorbox{green!30}{0.899}&\colorbox{green!30}{0.909}&\colorbox{green!30}{0.727}&\colorbox{green!30}{0.846}&\colorbox{green!30}{0.850}&\colorbox{green!30}{0.672}\\
BLeSS-FSIM \cite{temel2016bless}&\colorbox{green!30}{0.927}&\colorbox{green!30}{0.924}&\colorbox{green!30}{0.751}&\colorbox{green!30}{0.853}&\colorbox{green!30}{0.851}&\colorbox{green!30}{0.669}  \\
BLeSS-FSIMc \cite{temel2016bless}&\colorbox{green!30}{0.932}&\colorbox{green!30}{0.935}&\colorbox{green!30}{0.768}&\colorbox{green!30}{0.871}&\colorbox{green!30}{0.871}&\colorbox{green!30}{0.687}\\
LCSIM1 \cite{oszust2016full} & - & - & - &\colorbox{green!30}{0.896}&\colorbox{green!30}{0.906}&\colorbox{green!30}{0.727}\\
ReSIFT \cite{temel2016resift}&\colorbox{green!30}{0.914}&\colorbox{green!30}{0.917}&\colorbox{green!30}{0.733}&\colorbox{green!30}{0.627}&\colorbox{green!30}{0.632}&\colorbox{green!30}{0.484} \\
IQ($\mathcal{L}_T$) \cite{ali2017image}&\colorbox{green!30}{0.910}&\colorbox{green!30}{0.912}&\colorbox{green!30}{0.718}  &\colorbox{green!30}{0.841}&\colorbox{green!30}{0.840}&\colorbox{green!30}{0.629}\\
MS-UNIQUE \cite{prabhushankar2017ms}&\colorbox{green!30}{0.954}&\colorbox{green!30}{0.956} &\colorbox{green!30}{0.840}&\colorbox{green!30}{0.846}&\colorbox{green!30}{0.869}&\colorbox{green!30}{0.681} \\
SSIM CNN \cite{amirshahi2018reviving}&\colorbox{green!30}{0.917}&\colorbox{green!30}{0.921}&\colorbox{green!30}{0.743}&\colorbox{green!30}{0.770}&\colorbox{green!30}{0.737}&\colorbox{green!30}{0.551}\\
RVSIM \cite{yang2018rvsim}&\colorbox{green!30}{0.894}&\colorbox{green!30}{0.901}&\colorbox{green!30}{0.719}&\colorbox{green!30}{0.789}&\colorbox{green!30}{0.743}&\colorbox{green!30}{0.566}\\
2stepQA \cite{yu2019predicting}& \colorbox{green!30}{0.883}& \colorbox{green!30}{0.887}& \colorbox{green!30}{0.698}&\colorbox{green!30}{0.757}&\colorbox{green!30}{0.769}&\colorbox{green!30}{0.574}\\
SUMMER \cite{temel2019perceptual}&\colorbox{green!30}{0.750}&\colorbox{green!30}{0.754}& \colorbox{green!30}{0.596}&\colorbox{green!30}{0.817}&\colorbox{green!30}{0.823}&\colorbox{green!30}{0.637}  \\
CEQI \cite{layek2019center}&\colorbox{green!30}{0.894}&\colorbox{green!30}{0.920}&\colorbox{green!30}{0.747}&\colorbox{green!30}{0.887}&\colorbox{green!30}{0.891}&\colorbox{green!30}{0.714} \\
CEQIc \cite{layek2019center}&\colorbox{green!30}{0.906}&\colorbox{green!30}{0.918}&\colorbox{green!30}{0.744}&\colorbox{green!30}{0.892}&\colorbox{green!30}{0.895}&\colorbox{green!30}{0.719} \\
VCGS \cite{shi2020full}&\colorbox{green!30}{0.940}&\colorbox{green!30}{0.937}&\colorbox{green!30}{0.773}&\colorbox{green!30}{0.878}&\colorbox{green!30}{0.887}&\colorbox{green!30}{0.705}\\
DISTS \cite{ding2020image}&\colorbox{green!30}{0.923}&\colorbox{green!30}{0.922}&\colorbox{green!30}{0.746}&\colorbox{green!30}{0.705}&\colorbox{green!30}{0.668}&\colorbox{green!30}{0.488} \\
%SSIM CNN \cite{amirshahi2018reviving}& 0.924 &0.931  &0.800 \\
%DeepQA \cite{kim2017deep}&0.921/0.920 $(\pm0.008)$ &0.917/0.916 $(\pm0.007)$ &0.790/0.791 $(\pm0.009)$ \\
%WaDIQaM-FR \cite{bosse2017deep}&0.915/0.915 $(\pm0.006)$ & 0.913/0.912 $(\pm0.007)$ &0.784/0.784 $(\pm0.011)$ \\
\hline
DeepFL-IQA \cite{lin2020deepfl}&- &- &- &- &- &- \\
BLINDER \cite{gao2018blind}   &- &- &- &- &- &- \\
RankIQA \cite{liu2017rankiqa}   &- &- &- &- &- &- \\
BPSOM-MD \cite{pan2018blind}  &- &- &- &- &- &- \\
NSSADNN \cite{yan2019naturalness}   &- &- &- &- &- &- \\
\midrule
 \textit{ ActMapFeat} (ours)  & \textbf{0.960}  & \textbf{0.961} & \textbf{0.826}
 &\textbf{0.941} &\textbf{0.937} &\textbf{0.790}\\
\bottomrule
\end{tabular}  
\end{center}
\label{table:iqacomp2a}
\end{table}

\begin{table}
\caption{Performance comparison on MDID \cite{sun2017mdid}
and CSIQ \cite{larson2010most}
databases.
Mean PLCC, SROCC, and KROCC values
are reported for the learning-based approaches
measured over 100 random train-test splits.
The best results are typed in \textbf{bold}.
The
\colorbox{green!30}{green} background color stands for the fact that the correlation is
lower than those of the proposed method and the difference is statistically
significant with \emph{p} < 0.05. We used '-' if the data were not available.} %title of the table
\centering % centering table
\begin{center}
  \begin{tabular}{c  c  c  cccc}
\toprule
 & \multicolumn{3}{c  }{\textbf{ MDID \cite{sun2017mdid}}}&
 \multicolumn{3}{c  }{\textbf{ CSIQ \cite{larson2010most}}}\\%  & 
% \multicolumn{3}{c}{\textbf{ TID2013 \cite{ponomarenko2015image}}} \\
\cline{2-7}
 & \textbf{PLCC} & \textbf{SROCC} & \textbf{KROCC}& \textbf{PLCC} & \textbf{SROCC} & \textbf{KROCC} \\
\hline
% PSNR} &  0.653} &  0.653} &  0.518} & 0.647} & 0.647} & 0.488} \\
SSIM \cite{wang2004image}&\colorbox{green!30}{0.581}&\colorbox{green!30}{0.576}&\colorbox{green!30}{0.411}&\colorbox{green!30}{0.812}&\colorbox{green!30}{0.812}&\colorbox{green!30}{0.606} \\
MS-SSIM \cite{wang2003multiscale}&\colorbox{green!30}{0.836}&\colorbox{green!30}{0.841}&\colorbox{green!30}{0.654}&\colorbox{green!30}{0.913}&\colorbox{green!30}{0.917}&\colorbox{green!30}{0.743} \\
MAD \cite{larson2010most}&\colorbox{green!30}{0.742} &\colorbox{green!30}{0.725} &\colorbox{green!30}{0.533}&\colorbox{green!30}{0.950}&\colorbox{green!30}{0.947}&\colorbox{green!30}{0.796} \\
GSM \cite{liu2011image}&\colorbox{green!30}{0.825} &\colorbox{green!30}{0.827} &\colorbox{green!30}{0.636}&\colorbox{green!30}{0.906}&\colorbox{green!30}{0.910}&\colorbox{green!30}{0.729} \\
HaarPSI \cite{reisenhofer2018haar}&\colorbox{green!30}{0.904}&\colorbox{green!30}{0.903}&\colorbox{green!30}{0.734}&\colorbox{green!30}{0.946}&\colorbox{green!30}{0.960}&\colorbox{green!30}{0.823}\\
MDSI \cite{nafchi2016mean}&\colorbox{green!30}{0.829}&\colorbox{green!30}{0.836}&\colorbox{green!30}{0.653}&\colorbox{green!30}{0.953}&\colorbox{green!30}{0.957}&\colorbox{green!30}{0.812} \\
CSV \cite{temel2016csv}&\colorbox{green!30}{0.879}&\colorbox{green!30}{0.881}&\colorbox{green!30}{0.700}&\colorbox{green!30}{0.933}&\colorbox{green!30}{0.933}&\colorbox{green!30}{0.766}\\
GMSD \cite{xue2013gradient}&\colorbox{green!30}{0.864}&\colorbox{green!30}{0.862}&\colorbox{green!30}{0.680}&\colorbox{green!30}{0.954}&\colorbox{green!30}{0.957}&\colorbox{green!30}{0.812}\\
DSS \cite{balanov2015image}&\colorbox{green!30}{0.870}&\colorbox{green!30}{0.866}&\colorbox{green!30}{0.679}&\colorbox{green!30}{0.953}&\colorbox{green!30}{0.955}&\colorbox{green!30}{0.811} \\
VSI \cite{zhang2014vsi}&\colorbox{green!30}{0.855}&\colorbox{green!30}{0.857}&\colorbox{green!30}{0.671}&\colorbox{green!30}{0.928}&\colorbox{green!30}{0.942}&\colorbox{green!30}{0.785}\\
PerSIM \cite{temel2015persim}&\colorbox{green!30}{0.823}&\colorbox{green!30}{0.820}&\colorbox{green!30}{0.630}&\colorbox{green!30}{0.924}&\colorbox{green!30}{0.929}&\colorbox{green!30}{0.768}\\
BLeSS-SR-SIM \cite{temel2016bless}&\colorbox{green!30}{0.805}&\colorbox{green!30}{0.815}&\colorbox{green!30}{0.626}&\colorbox{green!30}{0.892}&\colorbox{green!30}{0.893}&\colorbox{green!30}{0.718}\\
BLeSS-FSIM \cite{temel2016bless}&\colorbox{green!30}{0.848}&\colorbox{green!30}{0.847}&\colorbox{green!30}{0.658}&\colorbox{green!30}{0.882}&\colorbox{green!30}{0.885}&\colorbox{green!30}{0.701}\\
BLeSS-FSIMc \cite{temel2016bless}&\colorbox{green!30}{0.878}&\colorbox{green!30}{0.883}&\colorbox{green!30}{0.702}&\colorbox{green!30}{0.913}&\colorbox{green!30}{0.917}&\colorbox{green!30}{0.743}\\
LCSIM1 \cite{oszust2016full}&-&-&-&\colorbox{green!30}{0.897}&\colorbox{green!30}{0.949}&\colorbox{green!30}{0.799}\\
ReSIFT \cite{temel2016resift}&\colorbox{green!30}{0.905}&\colorbox{green!30}{0.895}&\colorbox{green!30}{0.716}&\colorbox{green!30}{0.884}&\colorbox{green!30}{0.868}&\colorbox{green!30}{0.695}\\
IQ($\mathcal{L}_T$) \cite{ali2017image} &\colorbox{green!30}{0.867}&\colorbox{green!30}{0.865}&\colorbox{green!30}{0.708} & \colorbox{green!30}{0.915} & \colorbox{green!30}{0.912} & \colorbox{green!30}{0.720} \\
MS-UNIQUE \cite{prabhushankar2017ms}&\colorbox{green!30}{0.863}&\colorbox{green!30}{0.871}&\colorbox{green!30}{0.689}&\colorbox{green!30}{0.918}&\colorbox{green!30}{0.929}&\colorbox{green!30}{0.759}\\
SSIM CNN \cite{amirshahi2018reviving}&\colorbox{green!30}{0.904}&\colorbox{green!30}{0.907}&\colorbox{green!30}{0.732}&\colorbox{green!30}{0.952}&\colorbox{green!30}{0.946}&\colorbox{green!30}{0.794}\\
RVSIM \cite{yang2018rvsim}&\colorbox{green!30}{0.884}&\colorbox{green!30}{0.884}&\colorbox{green!30}{0.709}&\colorbox{green!30}{0.923}&\colorbox{green!30}{0.903}&\colorbox{green!30}{0.728}\\
2stepQA \cite{yu2019predicting}& \colorbox{green!30}{0.753}& \colorbox{green!30}{0.759}& \colorbox{green!30}{0.562}&\colorbox{green!30}{0.841}&\colorbox{green!30}{0.849}&\colorbox{green!30}{0.655}\\
SUMMER \cite{temel2019perceptual}&\colorbox{green!30}{0.742}&\colorbox{green!30}{0.734}&\colorbox{green!30}{0.543}&\colorbox{green!30}{0.826}&\colorbox{green!30}{0.830}&\colorbox{green!30}{0.658}\\
CEQI \cite{layek2019center}&\colorbox{green!30}{0.863}&\colorbox{green!30}{0.864}&\colorbox{green!30}{0.685}&\colorbox{green!30}{0.956}&\colorbox{green!30}{0.956}&\colorbox{green!30}{0.814} \\
CEQIc \cite{layek2019center}&\colorbox{green!30}{0.864}&\colorbox{green!30}{0.863}&\colorbox{green!30}{0.684}&\colorbox{green!30}{0.956}&\colorbox{green!30}{0.955}&\colorbox{green!30}{0.810}\\
VCGS \cite{shi2020full}&\colorbox{green!30}{0.867}&\colorbox{green!30}{0.869}&\colorbox{green!30}{0.687}&\colorbox{green!30}{0.931}&\colorbox{green!30}{0.944}&\colorbox{green!30}{0.790}\\
DISTS \cite{ding2020image}&\colorbox{green!30}{0.862}&\colorbox{green!30}{0.860}&\colorbox{green!30}{0.669}&\colorbox{green!30}{0.930}&\colorbox{green!30}{0.930}&\colorbox{green!30}{0.764}\\
%SSIM CNN \cite{amirshahi2018reviving}& 0.871 & 0.880 & 0.699\\
%DeepQA \cite{kim2017deep}& 0.955/0.955 $(\pm0.007)$& 0.954/0.953 $(\pm0.008)$& 0.810/0.811 $(\pm0.008)$\\
%WaDIQaM-FR \cite{bosse2017deep}& 0.949/0.948 $(\pm0.006)$ & 0.940/0.940 $(\pm0.008)$ & 0.802/0.803 $(\pm0.008)$\\
\hline
DeepFL-IQA \cite{lin2020deepfl} &- &- &- &0.946 &0.930 &- \\
BLINDER \cite{gao2018blind}   &- &- &- &0.968 &0.961 &- \\
RankIQA \cite{liu2017rankiqa}   &- &- &- &0.960 &0.947 &- \\
BPSOM-MD \cite{pan2018blind}  &- &- &- &0.860 &0.904 &- \\
NSSADNN \cite{yan2019naturalness}   &- &- &- &0.927 &0.893 &- \\
\midrule
\textit{ ActMapFeat}(ours)&\textbf{0.930}&\textbf{0.927}&\textbf{0.769}&\textbf{0.971}&\textbf{0.970}&\textbf{0.850}\\
\bottomrule
\end{tabular}  
\end{center}
\label{table:iqacomp4}
\end{table}

\unskip

\subsection{Cross Database Test}
\label{sec:cross}
\begin{figure}
\centering
\includegraphics[width=11cm]{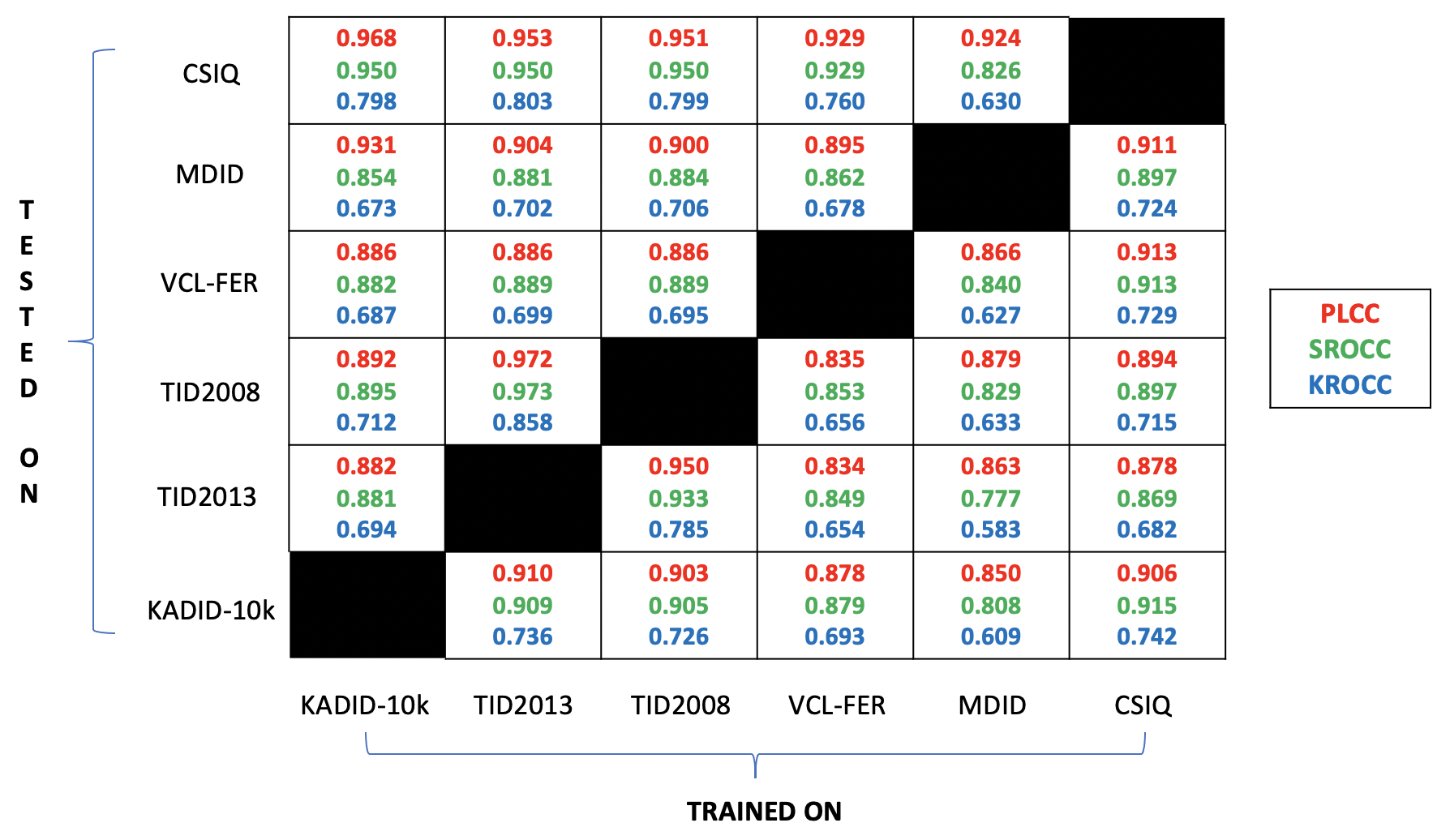}
\caption{Results of the cross database test in matrix form.
The proposed \textit{ActMapFeat} was trained on one given
IQA benchmark database (horizontal edge) and tested on another one (vertical edge).}
\label{fig:cross}
\end{figure}

Cross database test refers to the procedure of training on one given IQA benchmark
database and testing on another to show the generalization potential
of a machine learning based method. The~results of the cross database test
using KADID-10k \cite{lin2019kadid},
TID2013 \cite{ponomarenko2015image},
TID2008 \cite{ponomarenko2009tid2008},
VCL-FER \cite{zaric2012vcl},
MDID \cite{sun2017mdid}, and
CSIQ \cite{larson2010most} are depicted in Figure \ref{fig:cross}.
From these results, it can be concluded
that the proposed method loses
from its performance significantly in most
cases, but is still able to achieve
the performance of traditional 
state-of-the-art FR-IQA metrics.
Moreover, there are 
some pairings, such as
trained on KADID-10k \cite{lin2019kadid} and
tested on CSIQ \cite{larson2010most}, trained on
TID2013 \cite{ponomarenko2015image} and tested
on TID2008 \mbox{\cite{ponomarenko2009tid2008},} trained
on TID2008 \cite{ponomarenko2009tid2008}
and tested on TID2013 \cite{ponomarenko2015image}, where the performance loss is rather minor.

\section{Conclusions}
\label{sec:conc}
In this paper, we introduced a framework for FR-IQA relying on feature vectors,
which were obtained by comparing
reference and distorted activation maps by traditional image similarity metrics.
Unlike previous CNN-based approaches,
our method does not take patches from the distorted-reference image pairs, but
instead obtains convolutional activation
maps and creates feature vectors from these maps. This way, our method can be
easily generalized to any input image resolutions
or base CNN architecture. Furthermore, we carried out a detailed parameter
study with respect to the applied regression technique
and image similarity metric. Moreover, we pointed out that the proposed
feature extraction method is effective,
since our method is able to reach high PLCC, SROCC, and KROCC values trained
only on 5\% of KADID-10k images. Our algorithm
was compared to 26 other state-of-the-art FR-IQA methods on six benchmark
IQA databases, such as KADID-10k, TID2013, VCL-FER, MDID, TID2008, and CSIQ.
Our method was able to outperform
the state-of-the-art in terms of PLCC, SROCC, and KROCC, as well. 
The generalization ability of the proposed method
was confirmed in cross database tests.

To facilitate the reproducibility of the presented results, the MATLAB source
code of the introduced method and test environments
is available at \url{https://github.com/Skythianos/FRIQA-ActMapFeat}.

\bibliographystyle{unsrt} 
\bibliography{references}

\begin{thebibliography}{10}

\bibitem{lecun1998gradient}
Yann LeCun, L{\'e}on Bottou, Yoshua Bengio, and Patrick Haffner.
\newblock Gradient-based learning applied to document recognition.
\newblock {\em Proceedings of the IEEE}, 86(11):2278--2324, 1998.

\bibitem{szegedy2015going}
Christian Szegedy, Wei Liu, Yangqing Jia, Pierre Sermanet, Scott Reed, Dragomir
  Anguelov, Dumitru Erhan, Vincent Vanhoucke, and Andrew Rabinovich.
\newblock Going deeper with convolutions.
\newblock In {\em Proceedings of the IEEE conference on computer vision and
  pattern recognition}, pages 1--9, 2015.

\bibitem{oord2016wavenet}
Aaron van~den Oord, Sander Dieleman, Heiga Zen, Karen Simonyan, Oriol Vinyals,
  Alex Graves, Nal Kalchbrenner, Andrew Senior, and Koray Kavukcuoglu.
\newblock Wavenet: A generative model for raw audio.
\newblock {\em arXiv preprint arXiv:1609.03499}, 2016.

\bibitem{krizhevsky2012imagenet}
Alex Krizhevsky, Ilya Sutskever, and Geoffrey~E Hinton.
\newblock Imagenet classification with deep convolutional neural networks.
\newblock In {\em Advances in neural information processing systems}, pages
  1097--1105, 2012.

\bibitem{russakovsky2015imagenet}
Olga Russakovsky, Jia Deng, Hao Su, Jonathan Krause, Sanjeev Satheesh, Sean Ma,
  Zhiheng Huang, Andrej Karpathy, Aditya Khosla, Michael Bernstein, et~al.
\newblock Imagenet large scale visual recognition challenge.
\newblock {\em International journal of computer vision}, 115(3):211--252,
  2015.

\bibitem{sharif2014cnn}
Ali Sharif~Razavian, Hossein Azizpour, Josephine Sullivan, and Stefan Carlsson.
\newblock Cnn features off-the-shelf: an astounding baseline for recognition.
\newblock In {\em Proceedings of the IEEE conference on computer vision and
  pattern recognition workshops}, pages 806--813, 2014.

\bibitem{penatti2015deep}
Ot{\'a}vio~AB Penatti, Keiller Nogueira, and Jefersson~A Dos~Santos.
\newblock Do deep features generalize from everyday objects to remote sensing
  and aerial scenes domains?
\newblock In {\em Proceedings of the IEEE conference on computer vision and
  pattern recognition workshops}, pages 44--51, 2015.

\bibitem{bousetouane2015off}
Fouad Bousetouane and Brendan Morris.
\newblock Off-the-shelf cnn features for fine-grained classification of vessels
  in a maritime environment.
\newblock In {\em International Symposium on Visual Computing}, pages 379--388.
  Springer, 2015.

\bibitem{chou1995perceptually}
Chun-Hsien Chou and Yun-Chin Li.
\newblock A perceptually tuned subband image coder based on the measure of
  just-noticeable-distortion profile.
\newblock {\em IEEE Transactions on circuits and systems for video technology},
  5(6):467--476, 1995.

\bibitem{daly1992visible}
Scott~J Daly.
\newblock Visible differences predictor: an algorithm for the assessment of
  image fidelity.
\newblock In {\em Human Vision, Visual Processing, and Digital Display III},
  volume 1666, pages 2--15. International Society for Optics and Photonics,
  1992.

\bibitem{watson1997image}
Andrew~B Watson, Robert Borthwick, and Mathias Taylor.
\newblock Image quality and entropy masking.
\newblock In {\em Human vision and electronic imaging II}, volume 3016, pages
  2--12. International Society for Optics and Photonics, 1997.

\bibitem{wang2004image}
Zhou Wang, Alan~C Bovik, Hamid~R Sheikh, Eero~P Simoncelli, et~al.
\newblock Image quality assessment: from error visibility to structural
  similarity.
\newblock {\em IEEE transactions on image processing}, 13(4):600--612, 2004.

\bibitem{chen2006edge}
Guan-Hao Chen, Chun-Ling Yang, Lai-Man Po, and Sheng-Li Xie.
\newblock Edge-based structural similarity for image quality assessment.
\newblock In {\em 2006 IEEE International Conference on Acoustics Speech and
  Signal Processing Proceedings}, volume~2, pages II--II. IEEE, 2006.

\bibitem{wang2003multiscale}
Zhou Wang, Eero~P Simoncelli, and Alan~C Bovik.
\newblock Multiscale structural similarity for image quality assessment.
\newblock In {\em The Thrity-Seventh Asilomar Conference on Signals, Systems \&
  Computers, 2003}, volume~2, pages 1398--1402. Ieee, 2003.

\bibitem{li2009three}
Chaofeng Li and Alan~C Bovik.
\newblock Three-component weighted structural similarity index.
\newblock In {\em Image quality and system performance VI}, volume 7242, page
  72420Q. International Society for Optics and Photonics, 2009.

\bibitem{liu2011visual}
Hantao Liu and Ingrid Heynderickx.
\newblock Visual attention in objective image quality assessment: Based on
  eye-tracking data.
\newblock {\em IEEE Transactions on Circuits and Systems for Video Technology},
  21(7):971--982, 2011.

\bibitem{wang2010information}
Zhou Wang and Qiang Li.
\newblock Information content weighting for perceptual image quality
  assessment.
\newblock {\em IEEE Transactions on Image Processing}, 20(5):1185--1198, 2010.

\bibitem{zhang2011fsim}
Lin Zhang, Lei Zhang, Xuanqin Mou, and David Zhang.
\newblock Fsim: A feature similarity index for image quality assessment.
\newblock {\em IEEE transactions on Image Processing}, 20(8):2378--2386, 2011.

\bibitem{xue2013gradient}
Wufeng Xue, Lei Zhang, Xuanqin Mou, and Alan~C Bovik.
\newblock Gradient magnitude similarity deviation: A highly efficient
  perceptual image quality index.
\newblock {\em IEEE Transactions on Image Processing}, 23(2):684--695, 2013.

\bibitem{reisenhofer2018haar}
Rafael Reisenhofer, Sebastian Bosse, Gitta Kutyniok, and Thomas Wiegand.
\newblock A haar wavelet-based perceptual similarity index for image quality
  assessment.
\newblock {\em Signal Processing: Image Communication}, 61:33--43, 2018.

\bibitem{wang2008color}
Yuqing Wang, Weiya Liu, and Yong Wang.
\newblock Color image quality assessment based on quaternion singular value
  decomposition.
\newblock In {\em 2008 Congress on Image and Signal Processing}, volume~3,
  pages 433--439. IEEE, 2008.

\bibitem{kolaman2011quaternion}
Amir Kolaman and Orly Yadid-Pecht.
\newblock Quaternion structural similarity: a new quality index for color
  images.
\newblock {\em IEEE Transactions on Image Processing}, 21(4):1526--1536, 2011.

\bibitem{glowacz2010automated}
Andrzej G{\l}owacz, Micha{\l} Grega, Przemys{\l}aw Gwiazda, Lucjan Janowski,
  Miko{\l}aj Leszczuk, Piotr Romaniak, and Simon~Pietro Romano.
\newblock Automated qualitative assessment of multi-modal distortions in
  digital images based on glz.
\newblock {\em annals of telecommunications-annales des
  t{\'e}l{\'e}communications}, 65(1-2):3--17, 2010.

\bibitem{liang2016image}
Yudong Liang, Jinjun Wang, Xingyu Wan, Yihong Gong, and Nanning Zheng.
\newblock Image quality assessment using similar scene as reference.
\newblock In {\em European Conference on Computer Vision}, pages 3--18.
  Springer, 2016.

\bibitem{kim2017deep}
Jongyoo Kim and Sanghoon Lee.
\newblock Deep learning of human visual sensitivity in image quality assessment
  framework.
\newblock In {\em Proceedings of the IEEE conference on computer vision and
  pattern recognition}, pages 1676--1684, 2017.

\bibitem{sermanet2013overfeat}
Pierre Sermanet, David Eigen, Xiang Zhang, Micha{\"e}l Mathieu, Rob Fergus, and
  Yann LeCun.
\newblock Overfeat: Integrated recognition, localization and detection using
  convolutional networks.
\newblock {\em arXiv preprint arXiv:1312.6229}, 2013.

\bibitem{zhang2018unreasonable}
Richard Zhang, Phillip Isola, Alexei~A Efros, Eli Shechtman, and Oliver Wang.
\newblock The unreasonable effectiveness of deep features as a perceptual
  metric.
\newblock In {\em Proceedings of the IEEE Conference on Computer Vision and
  Pattern Recognition}, pages 586--595, 2018.

\bibitem{ali2017image}
Seyed Ali~Amirshahi, Marius Pedersen, and Stella~X Yu.
\newblock Image quality assessment by comparing cnn features between images.
\newblock {\em Electronic Imaging}, 2017(12):42--51, 2017.

\bibitem{bosse2017deep}
Sebastian Bosse, Dominique Maniry, Klaus-Robert M{\"u}ller, Thomas Wiegand, and
  Wojciech Samek.
\newblock Deep neural networks for no-reference and full-reference image
  quality assessment.
\newblock {\em IEEE Transactions on Image Processing}, 27(1):206--219, 2017.

\bibitem{simonyan2014very}
Karen Simonyan and Andrew Zisserman.
\newblock Very deep convolutional networks for large-scale image recognition.
\newblock {\em arXiv preprint arXiv:1409.1556}, 2014.

\bibitem{varga2020composition}
Domonkos Varga.
\newblock Composition-preserving deep approach to full-reference image quality
  assessment.
\newblock {\em Signal, Image and Video Processing}, pages 1--8, 2020.

\bibitem{okarma2010combined}
Krzysztof Okarma.
\newblock Combined full-reference image quality metric linearly correlated with
  subjective assessment.
\newblock In {\em International Conference on Artificial Intelligence and Soft
  Computing}, pages 539--546. Springer, 2010.

\bibitem{sheikh2006image}
Hamid~R Sheikh and Alan~C Bovik.
\newblock Image information and visual quality.
\newblock {\em IEEE Transactions on image processing}, 15(2):430--444, 2006.

\bibitem{mansouri2009image}
Azadeh Mansouri, Ahmad~Mahmoudi Aznaveh, Farah Torkamani-Azar, and J~Afshar
  Jahanshahi.
\newblock Image quality assessment using the singular value decomposition
  theorem.
\newblock {\em Optical Review}, 16(2):49--53, 2009.

\bibitem{okarma2012combined}
Krzysztof Okarma.
\newblock Combined image similarity index.
\newblock {\em Optical Review}, 19(5):349--354, 2012.

\bibitem{okarma2013extended}
K~Okarma.
\newblock Extended hybrid image similarity--combined full-reference image
  quality metric linearly correlated with subjective scores.
\newblock {\em Elektronika ir Elektrotechnika}, 19(10):129--132, 2013.

\bibitem{oszust2017image}
Mariusz Oszust.
\newblock Image quality assessment with lasso regression and pairwise score
  differences.
\newblock {\em Multimedia Tools and Applications}, 76(11):13255--13270, 2017.

\bibitem{yuan2015image}
Yuan Yuan, Qun Guo, and Xiaoqiang Lu.
\newblock Image quality assessment: a sparse learning way.
\newblock {\em Neurocomputing}, 159:227--241, 2015.

\bibitem{lukin2015combining}
Vladimir~V Lukin, Nikolay~N Ponomarenko, Oleg~I Ieremeiev, Karen~O Egiazarian,
  and Jaakko Astola.
\newblock Combining full-reference image visual quality metrics by neural
  network.
\newblock In {\em Human Vision and Electronic Imaging XX}, volume 9394, page
  93940K. International Society for Optics and Photonics, 2015.

\bibitem{oszust2016full}
Mariusz Oszust.
\newblock Full-reference image quality assessment with linear combination of
  genetically selected quality measures.
\newblock {\em PloS one}, 11(6):e0158333, 2016.

\bibitem{amirshahi2018reviving}
Seyed~Ali Amirshahi, Marius Pedersen, and Azeddine Beghdadi.
\newblock Reviving traditional image quality metrics using cnns.
\newblock In {\em Color and Imaging Conference}, volume 2018, pages 241--246.
  Society for Imaging Science and Technology, 2018.

\bibitem{lin2020deepfl}
Hanhe Lin, Vlad Hosu, and Dietmar Saupe.
\newblock Deepfl-iqa: Weak supervision for deep iqa feature learning.
\newblock {\em arXiv preprint arXiv:2001.08113}, 2020.

\bibitem{gao2018blind}
Fei Gao, Jun Yu, Suguo Zhu, Qingming Huang, and Qi~Tian.
\newblock Blind image quality prediction by exploiting multi-level deep
  representations.
\newblock {\em Pattern Recognition}, 81:432--442, 2018.

\bibitem{lin2019kadid}
Hanhe Lin, Vlad Hosu, and Dietmar Saupe.
\newblock Kadid-10k: A large-scale artificially distorted iqa database.
\newblock In {\em 2019 Eleventh International Conference on Quality of
  Multimedia Experience (QoMEX)}, pages 1--3. IEEE, 2019.

\bibitem{lin2018koniq}
Hanhe Lin, Vlad Hosu, and Dietmar Saupe.
\newblock Koniq-10k: Towards an ecologically valid and large-scale iqa
  database.
\newblock {\em arXiv preprint arXiv:1803.08489}, 2018.

\bibitem{ponomarenko2009tid2008}
Nikolay Ponomarenko, Vladimir Lukin, Alexander Zelensky, Karen Egiazarian,
  Marco Carli, and Federica Battisti.
\newblock Tid2008-a database for evaluation of full-reference visual quality
  assessment metrics.
\newblock {\em Advances of Modern Radioelectronics}, 10(4):30--45, 2009.

\bibitem{zaric2012vcl}
An{\dj}ela Zari{\'c}, Nenad Tatalovi{\'c}, Nikolina Brajkovi{\'c}, Hrvoje
  Hlevnjak, Matej Lon{\v{c}}ari{\'c}, Emil Dumi{\'c}, and Sonja Grgi{\'c}.
\newblock Vcl@ fer image quality assessment database.
\newblock {\em AUTOMATIKA: {\v{c}}asopis za automatiku, mjerenje, elektroniku,
  ra{\v{c}}unarstvo i komunikacije}, 53(4):344--354, 2012.

\bibitem{sun2017mdid}
Wen Sun, Fei Zhou, and Qingmin Liao.
\newblock Mdid: A multiply distorted image database for image quality
  assessment.
\newblock {\em Pattern Recognition}, 61:153--168, 2017.

\bibitem{larson2010most}
Eric~Cooper Larson and Damon~Michael Chandler.
\newblock Most apparent distortion: full-reference image quality assessment and
  the role of strategy.
\newblock {\em Journal of electronic imaging}, 19(1):011006, 2010.

\bibitem{hii2017multigap}
Yong-Lian Hii, John See, Magzhan Kairanbay, and Lai-Kuan Wong.
\newblock Multigap: Multi-pooled inception network with text augmentation for
  aesthetic prediction of photographs.
\newblock In {\em 2017 IEEE International Conference on Image Processing
  (ICIP)}, pages 1722--1726. IEEE, 2017.

\bibitem{drucker1997support}
Harris Drucker, Christopher~JC Burges, Linda Kaufman, Alex~J Smola, and
  Vladimir Vapnik.
\newblock Support vector regression machines.
\newblock In {\em Advances in neural information processing systems}, pages
  155--161, 1997.

\bibitem{williams2006gaussian}
Christopher~KI Williams and Carl~Edward Rasmussen.
\newblock {\em Gaussian processes for machine learning}, volume~2.
\newblock MIT press Cambridge, MA, 2006.

\bibitem{ponomarenko2015image}
Nikolay Ponomarenko, Lina Jin, Oleg Ieremeiev, Vladimir Lukin, Karen
  Egiazarian, Jaakko Astola, Benoit Vozel, Kacem Chehdi, Marco Carli, Federica
  Battisti, et~al.
\newblock Image database tid2013: Peculiarities, results and perspectives.
\newblock {\em Signal Processing: Image Communication}, 30:57--77, 2015.

\bibitem{sheikh2006statistical}
Hamid~R Sheikh, Muhammad~F Sabir, and Alan~C Bovik.
\newblock A statistical evaluation of recent full reference image quality
  assessment algorithms.
\newblock {\em IEEE Transactions on image processing}, 15(11):3440--3451, 2006.

\bibitem{cho2015much}
Junghwan Cho, Kyewook Lee, Ellie Shin, Garry Choy, and Synho Do.
\newblock How much data is needed to train a medical image deep learning system
  to achieve necessary high accuracy?
\newblock {\em arXiv preprint arXiv:1511.06348}, 2015.

\bibitem{liu2011image}
Anmin Liu, Weisi Lin, and Manish Narwaria.
\newblock Image quality assessment based on gradient similarity.
\newblock {\em IEEE Transactions on Image Processing}, 21(4):1500--1512, 2011.

\bibitem{nafchi2016mean}
Hossein~Ziaei Nafchi, Atena Shahkolaei, Rachid Hedjam, and Mohamed Cheriet.
\newblock Mean deviation similarity index: Efficient and reliable
  full-reference image quality evaluator.
\newblock {\em IEEE Access}, 4:5579--5590, 2016.

\bibitem{temel2016csv}
Dogancan Temel and Ghassan AlRegib.
\newblock Csv: Image quality assessment based on color, structure, and visual
  system.
\newblock {\em Signal Processing: Image Communication}, 48:92--103, 2016.

\bibitem{balanov2015image}
Amnon Balanov, Arik Schwartz, Yair Moshe, and Nimrod Peleg.
\newblock Image quality assessment based on dct subband similarity.
\newblock In {\em 2015 IEEE International Conference on Image Processing
  (ICIP)}, pages 2105--2109. IEEE, 2015.

\bibitem{zhang2014vsi}
Lin Zhang, Ying Shen, and Hongyu Li.
\newblock Vsi: A visual saliency-induced index for perceptual image quality
  assessment.
\newblock {\em IEEE Transactions on Image Processing}, 23(10):4270--4281, 2014.

\bibitem{temel2015persim}
Dogancan Temel and Ghassan AlRegib.
\newblock Persim: Multi-resolution image quality assessment in the perceptually
  uniform color domain.
\newblock In {\em 2015 IEEE International Conference on Image Processing
  (ICIP)}, pages 1682--1686. IEEE, 2015.

\bibitem{temel2016bless}
Dogancan Temel and Ghassan AlRegib.
\newblock Bless: Bio-inspired low-level spatiochromatic similarity assisted
  image quality assessment.
\newblock In {\em 2016 IEEE International Conference on Multimedia and Expo
  (ICME)}, pages 1--6. IEEE, 2016.

\bibitem{temel2016resift}
Dogancan Temel and Ghassan AlRegib.
\newblock Resift: Reliability-weighted sift-based image quality assessment.
\newblock In {\em 2016 IEEE International Conference on Image Processing
  (ICIP)}, pages 2047--2051. IEEE, 2016.

\bibitem{prabhushankar2017ms}
Mohit Prabhushankar, Dogancan Temel, and Ghassan AlRegib.
\newblock Ms-unique: Multi-model and sharpness-weighted unsupervised image
  quality estimation.
\newblock {\em Electronic Imaging}, 2017(12):30--35, 2017.

\bibitem{yang2018rvsim}
Guangyi Yang, Deshi Li, Fan Lu, Yue Liao, and Wen Yang.
\newblock Rvsim: a feature similarity method for full-reference image quality
  assessment.
\newblock {\em EURASIP Journal on Image and Video Processing}, 2018(1):6, 2018.

\bibitem{yu2019predicting}
Xiangxu Yu, Christos~G Bampis, Praful Gupta, and Alan~Conrad Bovik.
\newblock Predicting the quality of images compressed after distortion in two
  steps.
\newblock {\em IEEE Transactions on Image Processing}, 28(12):5757--5770, 2019.

\bibitem{temel2019perceptual}
Dogancan Temel and Ghassan AlRegib.
\newblock Perceptual image quality assessment through spectral analysis of
  error representations.
\newblock {\em Signal Processing: Image Communication}, 70:37--46, 2019.

\bibitem{layek2019center}
Md~Layek, AFM Uddin, Tuyen~P Le, TaeChoong Chung, Eui-Nam Huh, et~al.
\newblock Center-emphasized visual saliency and a contrast-based full reference
  image quality index.
\newblock {\em Symmetry}, 11(3):296, 2019.

\bibitem{shi2020full}
Chenyang Shi and Yandan Lin.
\newblock Full reference image quality assessment based on visual salience with
  color appearance and gradient similarity.
\newblock {\em IEEE Access}, 2020.

\bibitem{ding2020image}
Keyan Ding, Kede Ma, Shiqi Wang, and Eero~P Simoncelli.
\newblock Image quality assessment: Unifying structure and texture similarity.
\newblock {\em arXiv preprint arXiv:2004.07728}, 2020.

\bibitem{liu2017rankiqa}
Xialei Liu, Joost van~de Weijer, and Andrew~D Bagdanov.
\newblock Rankiqa: Learning from rankings for no-reference image quality
  assessment.
\newblock In {\em Proceedings of the IEEE International Conference on Computer
  Vision}, pages 1040--1049, 2017.

\bibitem{pan2018blind}
Da~Pan, Ping Shi, Ming Hou, Zefeng Ying, Sizhe Fu, and Yuan Zhang.
\newblock Blind predicting similar quality map for image quality assessment.
\newblock In {\em Proceedings of the IEEE Conference on Computer Vision and
  Pattern Recognition}, pages 6373--6382, 2018.

\bibitem{yan2019naturalness}
Bo~Yan, Bahetiyaer Bare, and Weimin Tan.
\newblock Naturalness-aware deep no-reference image quality assessment.
\newblock {\em IEEE Transactions on Multimedia}, 21(10):2603--2615, 2019.

\bibitem{itu20121401}
P~ITU-T.
\newblock 1401: Methods, metrics and procedures for statistical evaluation,
  qualification and comparison of objective quality prediction models.
\newblock {\em ITU-T Recommendation}, page 1401, 2012.

\end{thebibliography}
\end{document}